\pdfoutput=1
\documentclass[11pt]{article}

\usepackage[final]{acl}

\usepackage{times}
\usepackage{latexsym}

\usepackage[T1]{fontenc}
\usepackage[utf8]{inputenc}

\usepackage{microtype}

\usepackage{inconsolata}

\usepackage{graphicx}

\usepackage{multirow}
\usepackage{pifont} % For cross symbol
\usepackage{enumitem}
\usepackage{amsmath}
\usepackage{colortbl}
\usepackage{pgf}
\usepackage{pgfplots}
\usepackage{import}
\usepackage[inkscapelatex=false]{svg}
\usepackage{subcaption}
\usepackage{caption}

\newcommand{\procsrl}{\textit{Implicit-VidSRL}}
\newcommand{\procsrlmodel}{\textit{iSRL-Qwen2-VL}}
\newcommand{\procsrlmodelT}{\textit{iSRL-Qwen2}}

\newcommand{\quotes}[1]{``#1''}

\newcommand{\R}[1]{{%
    \textbf{%
        \ifstrequal{#1}{1}{\textcolor{red}{R#1}}{%
        \ifstrequal{#1}{2}{\textcolor{blue}{R#1}}{%
        \ifstrequal{#1}{3}{\textcolor{magenta}{R#1}}{%
        \ifstrequal{#1}{4}{\textcolor{teal}{R#1}}{%
                           \textcolor{cyan}{R#1}%
        }}}}%
    }%
}}

\newcommand{\tick}{\textcolor{blue}{\ding{51}}}
\newcommand{\cross}{\textcolor{red}{\ding{55}}}

\newcommand{\srl}[3]{
  \{$\textcolor{orange}{verb}\text{:}#1$, %
  $\textcolor{OliveGreen}{what}\text{:}#2$, %
  $\textcolor{Magenta}{where/with}\text{:}#3$\}%
}
\newcommand{\action}{\textcolor{orange}{\textit{verb}}}
\newcommand{\what}{\textcolor{OliveGreen}{\textit{what}}}%
\newcommand{\where}{\textcolor{Magenta}{\textit{where/with}}}%
\newcommand{\iwhat}{\textcolor{OliveGreen}{\textit{what-implicit}}}%
\newcommand{\iwhere}{\textcolor{Magenta}{\textit{where/with-implicit}}}

\title{Predicting Implicit Arguments in Procedural Video Instructions}

\author{
 \textbf{Anil Batra\textsuperscript{1}} \quad
 \textbf{Laura Sevilla-Lara\textsuperscript{1}} \quad
 \textbf{Marcus Rohrbach\textsuperscript{2}} \quad
 \textbf{Frank Keller\textsuperscript{1}}
\\
\\
 \textsuperscript{1}University of Edinburgh, United Kingdom \quad
 \textsuperscript{2}TU Darmstadt \& hessian.AI, Germany
\\
    \texttt{a.k.batra@sms.ed.ac.uk} \quad \texttt{l.sevilla@ed.ac.uk}
\\
 \texttt{marcus.rohrbach@tu-darmstadt.de} \quad \texttt{keller@inf.ed.ac.uk}
 \\
 \small{
   \textbf{Project Page:} \href{https://anilbatra2185.github.io/p/ividsrl/}{https://anilbatra2185.github.io/p/ividsrl/}
 }
}

\begin{document}
\maketitle

\begin{abstract}
Procedural texts help AI enhance reasoning about context and action sequences. Transforming these into Semantic Role Labeling (SRL) improves understanding of individual steps by identifying predicate-argument structure like \{\action,\what,\where\}. Procedural instructions are highly elliptic, for instance, (i) \textit{add cucumber to the bowl} and (ii) \textit{add sliced tomatoes}, the second step's \textit{where} argument is inferred from the context, referring to where the cucumber was placed. Prior SRL benchmarks often miss implicit arguments, leading to incomplete understanding. To address this, we introduce \procsrl, a dataset that necessitates inferring implicit and explicit arguments from contextual information in multimodal cooking procedures. Our proposed dataset benchmarks multimodal models' contextual reasoning, requiring entity tracking through visual changes in recipes. We study recent multimodal LLMs and reveal that they struggle to predict implicit arguments of \what~and \where~from multimodal procedural data given the \action. Lastly, we propose \procsrlmodel, which achieves a 17\% relative improvement in F1-score for \iwhat~and a 14.7\% for \iwhere~semantic roles over GPT-4o. Dataset and Code are publicly available\footnote{https://github.com/anilbatra2185/implicit-vid-srl.git}.
\end{abstract}
\section{Introduction}
\begin{figure}[t]
    \centering
    \includegraphics[width=0.95\linewidth]{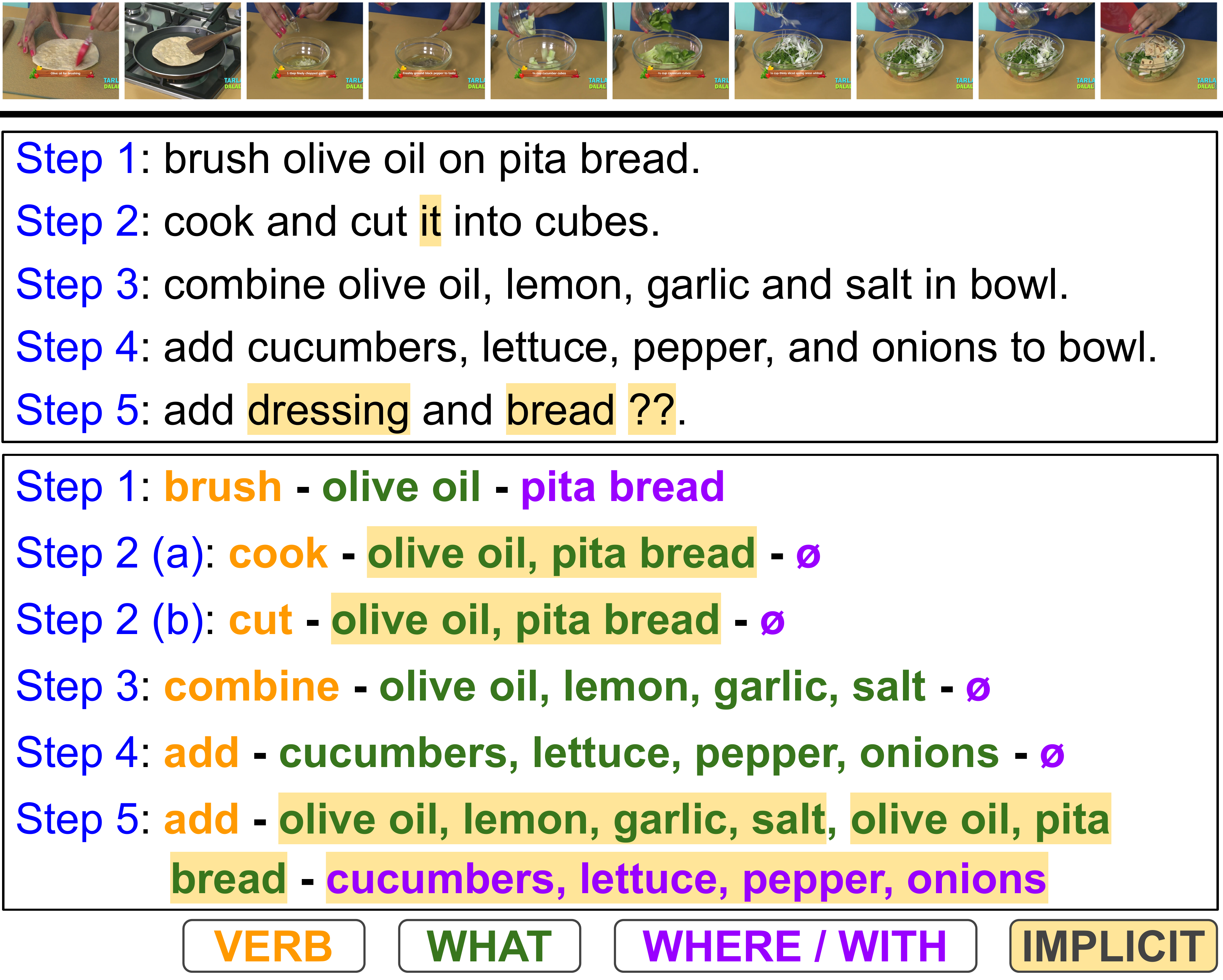}
    \caption{\textbf{\procsrl}: A new semantic role labeling (SRL) based dataset, to represent procedural videos using semantic frames (\{\action,\what,\where\}) with implict arguments. For instance \textit{step~2} is transformed into \textit{step~2(a)} \& \textit{step~2(b)}. While in step~5 the arguments are implicit and require both visual and textual context to infer from step~3 \&~2.
    The implicit information is emphasized using a background color.
    }
    \label{fig:teaser}
\end{figure}

When humans understand procedural videos, they typically rely on verbal instructions. For example, in cooking videos, the narrator will explain the individual recipe steps as they are carried out in the video. Crucially, these verbal instructions are often highly elliptic, as a lot of information can be inferred from the visual and linguistic context.

From Figure~\ref{fig:teaser}, take the instruction sequence  (1)~\textit{brush olive oil on pita bread} (2)~\textit{cook and cut it into cubes}. Here, we infer that in step~2, pita bread with oil is cooked and then the oiled pita bread after cooking is cut into cubes. This inference relies on linguistic context, including pronominal reference. In step~5 of Figure~\ref{fig:teaser}, we infer that \textit{dressing} refers to the mixture of \textit{olive oil, lemon, garlic, and salt} from step~3, while \textit{bread} refers to the cubes of oiled pita bread from step~2, which are later added to a bowl containing \textit{cucumbers, lettuce, pepper, and onions}. Linguistic context alone is insufficient to support this inference, but the visual context resolves the ambiguity -- we are dealing with three distinct mixtures: \textit{dressing, salad, bread}.

We need to capture such inferences in order to build models that understand complex, multi-step instructional videos. More specifically, our models need to predict \textbf{implicit arguments}, such as \textit{olive oil, pita bread} in step~2 and \textit{cucumbers, lettuce, \ldots} in step~5. In this work, we propose to use \textbf{semantic role labeling} (SRL) to model the semantics of narrated instructional video as simple predicate-argument structures. We show how this approach can be used to capture both explicit and implicit arguments in instructional steps, enabling more comprehensive video understanding and more informative system evaluation. We use a domain-specific variant of traditional SRL annotation, using \{\action, \what, \where\} tuples to represent semantic frames in multi-step videos (see Figure~\ref{fig:teaser}). Our focus is on implicit arguments that pertain to recipe ingredients in the \what~and \where~roles. We annotate a subset of videos from the standard procedural datasets i.e. YouCook2~\cite{ZhXuCoAAAI18} and Tasty~\cite{sener2022transferring}, resulting in multimodal procedural data with SRLs, the \procsrl~ dataset. We use a cloze task and a next-step prediction task to evaluate a model's comprehension of procedural data. The cloze task requires the model to predict the \what~and \where~roles (given the \action); this is a way of assessing the model's contextual reasoning skills. In next-step prediction, the model predicts the full semantic frame—including the verb and arguments (both implicit and explicit) — and generates the instruction text for the next step, using the previous steps as context. We evaluate SRL predictions with F1-scores for argument identification, verb recall for next-step prediction, and conventional natural language generation (NLG) metrics for generated text.

We assess various large multimodal models across both tasks and apply chain-of-thought (CoT) prompting~\cite{wei2022chain} to enhance the models' ability to predict implicit arguments. However, they still face difficulties in inferring implicit information from temporal contexts. To address this, we automatically create a training dataset with silver-standard SRL annotations using GPT-4o through in-context learning. We show that fine-tuning both the text and video versions of the Qwen2 model on silver-standard semantic labels enhances implicit argument prediction. Our \procsrlmodel~achieves 17\% relative F1-score improvement for \what~and 14.7\% for \where~over GPT-4o with multimodal input.

We summarize our \textbf{contributions} as: \textbf{(i)}~We present \procsrl, an SRL-based benchmark for the understanding of procedural steps in instructional videos, \textbf{(ii)}~We show that implicit argument prediction is challenging for multimodal models such as GPT-4o and Qwen2-VL, \textbf{(iii)}~Using our SRL scheme as an intermediate representation in the next step prediction task, we show that it boosts the Qwen2-VL model's performance in predicting future steps, leading to a $\sim$2\% improvement in the METEOR score, \textbf{(iv)}~We show that large multimodal models achieve good performance in next-step and implicit argument prediction when fine-tuned on silver-standard SRL annotations.
\section{Related work}
\paragraph{Semantic Role Labels} Semantic role labeling (SRL)  is a key NLP task that aids applications like document summarization~\cite{fan-etal-2023-evaluating}, building knowledge graphs~\cite{9356337}, machine translation~\cite{shi2016knowledge} and question answering~\cite{berant2013semantic}. The goal of SRL is to identify and label the arguments of the semantic predicates in a sentence, providing a shallow semantic representation. The process involves two stages: first, identifying the predicate (typically a verb), and second, determining its arguments along with their specific roles, such as \what~or \where. Previous linguistic benchmarks, such as PropBank~\cite{palmer2005proposition}, FrameNet~\cite{baker-etal-1998-berkeley-framenet}, NomBank~\cite{meyers2004nombank}, and the CoNLL SRL task \cite{carreras2005introduction}, concentrate on single sentences. These datasets mainly focus on explicit arguments; however, \citet{gerber2010beyond} expand NomBank by incorporating implicit information gathered from all preceding sentences in the document through manual annotation. The work focuses on a predefined set of 10~nominal predicates and shows that implicit arguments improve the role coverage of predicates, bridging the gap between human and machine understanding of text. In contrast, we focus on verb predicates in procedural text and annotate implicit arguments using all the entities mentioned in the previous context.

SRL has also been used for video understanding~\cite{yang2016grounded, sadhu2021visual}. \citet{yang2016grounded} extend the TACoS corpus~\cite{tacos:regnerietal:tacl} with bounding box annotations for verb predicate, along with explicit and implicit arguments. This prior research on implicit labels has limitations: (i)~it focuses on single short clips, (ii)~it only covers ``cutting cucumber'' and ``cutting bread'' tasks, and (iii)~the implicit information is local in text or video. Later, \citet{sadhu2021visual} introduced the VidSitu dataset for video-SRL, and \citet{khan2022grounded} extended it with argument bounding boxes. However, this prior work either focuses on local implicit information or explicit information in short temporal contexts. In contrast, we incorporate implicit information that requires the model to access an extended temporal context to identify ingredients related to the cooking actions. Moreover, in the cooking domain, the visual appearance and the composition of ingredients can change over time, which makes the task challenging.
\paragraph{Temporal Reasoning} As video-LLMs advance, several benchmarks have emerged to evaluate their reasoning abilities. SEED-Bench~\cite{Li_2024_CVPR} targets spatial or temporal comprehension, with a specific focus on the sequence of actions in procedural understanding. TempCompass~\cite{liu-etal-2024-tempcompass} analyzes short clips featuring a single object, concentrating on aspects like speed, direction, and the order of actions. SOKBench~\cite{wang2024sok} aligns with our interests as it uses videos-captions pairs from YouCook2~\cite{ZhXuCoAAAI18} to create graphs linking objects and entities. However, it often faces issues with underspecified instructions, leading to incomplete knowledge graphs and QA pairs. Our dataset, in contrast, offers comprehensive information based on which complete graphs and effective QA pairs could be generated. Current work focuses on open vocabulary generative predictions and we plan to introduce a QA-based reasoning benchmark in the future.
\paragraph{Procedural Step Anticipation} Predicting human actions has been extensively researched, enabling digital assistants to anticipate behaviors for improved support and protection. Leveraging the planning capabilities of LLMs, recent studies have utilized multimodal LLMs for next-step anticipation for long-term forecasting~\cite{zhao2023antgpt, islam2024propose}. Such efforts focus on coarse-grained action anticipation using \textit{verb-noun} pairs (e.g., \textit{take knife}), though some work~\cite{abdelsalam2023gepsan,sener2022transferring} explores fine-grained textual predictions. However, the limitations of natural language generation (NLG) metrics make evaluation difficult. In this work, we use additional metrics based on semantic frames to evaluate the fine-grained next step predictions. 
\paragraph{Procedural Understanding} Procedural learning is challenging with text-only or multimodal datasets, as it involves parsing recipe steps~\cite{yamakata2020english} and understanding visual dynamics~\cite{ZhXuCoAAAI18}. As part of recent work on the LLM-based understanding of procedural text, \citet{diallo2024pizzacommonsense} introduce \textit{PizzaCommonSense}. In this dataset, input-output pairs are annotated with intermediate step outputs. For instance, given two instructions, (i) `combine yeast, sugar, water', (ii) `stir gently to dissolve', the corresponding annotations are (i) `combine -- yeast, sugar, water -- yeast mixture', (ii) `stir -- yeast mixture -- yeast mixture'. Such annotations focus on explicit entities and lack an understanding of the compositions, e.g., `yeast mixture' is made with yeast, sugar and water. In contrast, our annotation contains implicit information and focuses on \{\action, \what, \where\}, i.e., (i) `combine -- yeast, sugar, water -- $\emptyset$', (ii) `stir -- yeast, sugar, water -- $\emptyset$'. Moreover, \textit{PizzaCommonSense} is restricted to pizza recipes and only includes text, whereas our dataset,~\procsrl, is multimodal and features a diverse range of cooking recipes.

In prior work, multi-modal instructional video dataset are explored for learning procedural knowledge~\cite{lin2022learning} and procedure planning~\cite{chang2020procedure}. Researchers have proposed various datasets for understanding instructional videos~\cite{Damen2021PAMI,zhukov2019cross, tang2019coin,Afouras_2023_htstep}, typically featuring annotations with fixed labels or brief step captions. In contrast, we focus on extended captions composed in natural English~\cite{ZhXuCoAAAI18,sener2022transferring}. It is also notable that cooking recipes make up a significant portion of existing datasets such as Epic-Kitchens~\cite{Damen2021PAMI} (100\%), CrossTask~\cite{zhukov2019cross} (82\%), HT-Step~\cite{Afouras_2023_htstep} (100\%), Exo-Ego4D~\cite{Grauman_2024_CVPR} (40\%), Guide~\cite{liang2024guide} (51\%). There are only a few datasets where cooking has an equal representation among other domains, such as COIN~\cite{tang2019coin}, or that concentrate on other domains, like Assembly-101~\cite{sener2022assembly101}. Cooking recipes are conceptually challenging, with many elliptical instructions and state transformations (shape or visual changes). These can be learned from large-scale data on platforms like YouTube and applied to other domains. Thus, we use multimodal cooking recipes, inspired by prior work and existing datasets.
\section{The \procsrl~Dataset}
We utilize the cooking videos from the validation and test sets of YouCook2~\cite{ZhXuCoAAAI18} and Tasty~\cite{sener2022transferring}. 
We have an untrimmed procedural video with multi-step instructions needed to complete a goal (e.g., a person following a cooking recipe). Each instruction is labeled with start and end times, delimiting the instruction temporally and a fine-grained description. Let $\mathcal{V}$ be a corpus of procedural videos with corresponding instructions and goals. Formally, each video is represented as $\mathbf{v} = (\mathcal{G}, \{(t_i^s, t_i^e, s_i)\}_{i=1}^{i=N})$, where $\mathcal{G}$ is title or goal of the video with $N$ instructions, $(t_i^s, t_i^e)$ are start and end time for the video clip, and $s_i$ represents a single multi-step instruction. We choose videos for the test set according to the criteria video duration ($\geq 30$ sec and $\leq 10$ min), number of instructions ($\geq 4$), valid YouTube videos, and presence of implicit arguments.

Following traditional SRL annotation schemes \citep{gildea2002automatic, do2017improving}, we propose to decompose a multi-action instruction $s_i$ paired with video clips into $M$ simple predicate-argument structures $\{a_j\}_{m=1}^M$. In our annotation scheme, each $a_j$ is the tuple of the form \{\action, \what, \where\}, where \action~is the primary action, \what~denotes the objects impacted by the action, and \where~refers to the location, context, or accompanying elements of the action. For example, step~2 in Figure~\ref{fig:teaser} is decomposed into two semantic frames, step~2(a)~\srl{cook}{[olive\ oil, pita\ bread]}{\emptyset} and step~2(b)~\srl{cut}{[olive\ oil, pita\ bread]}{\emptyset}, where $\emptyset$ signifies an empty argument. Note that the semantic frame can include implicit arguments, i.e., arguments of the verb that are inferred linguistically or visually. For example, in step 5, \what~in the semantic frame references the mixture from steps 3 and 2, which is combined with the vegetable mixture from step 4, needing both multimodal information. In this work, we focus only on implicit arguments for recipe ingredients, limited to the \what~and \where~roles.

\begin{figure*}[t]
    \centering
    \includegraphics[width=0.9\linewidth]{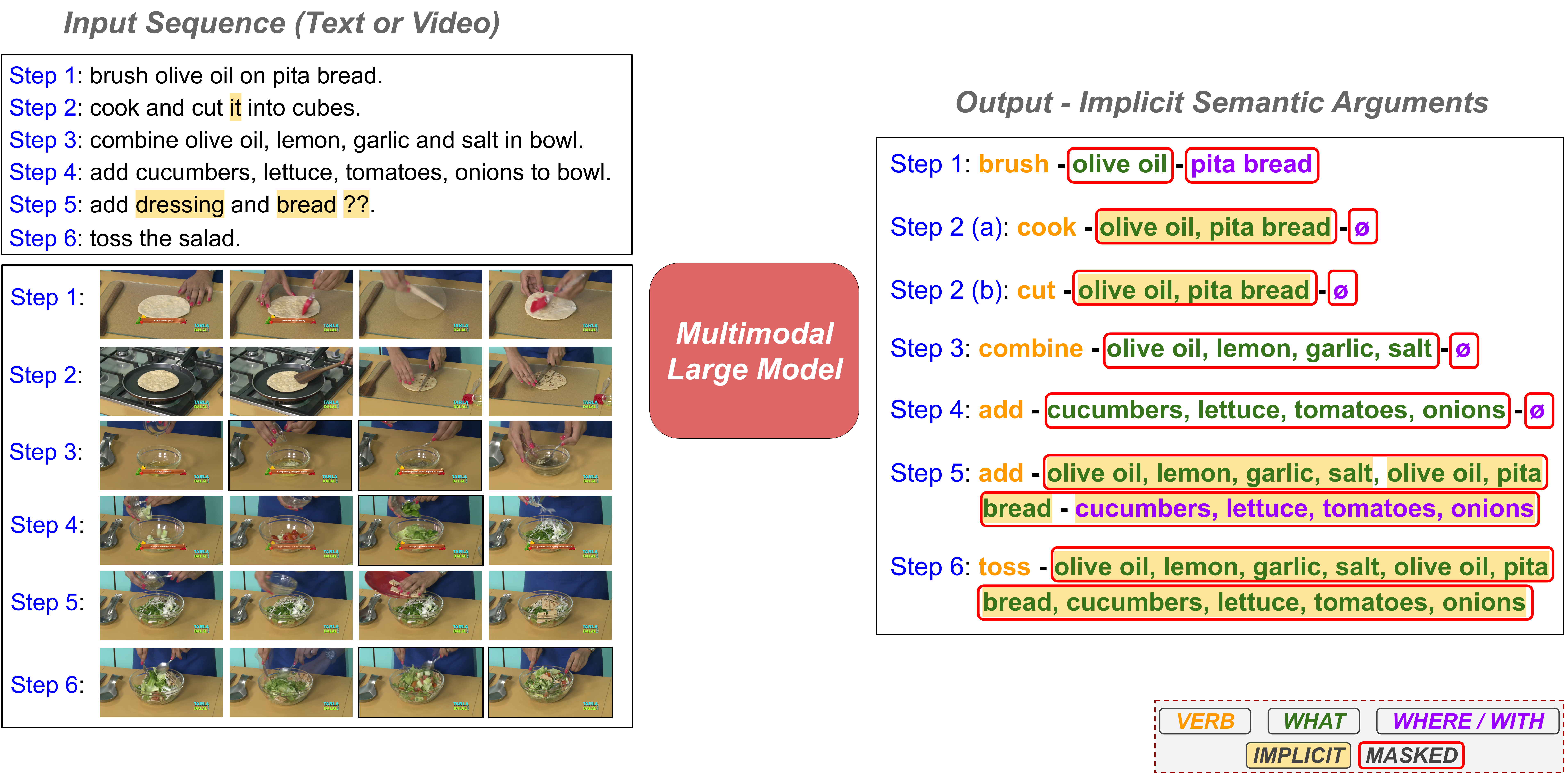}
    \caption{The \textit{Implicit Argument Prediction} task involves providing the input sequence, which may be in the form of text or video or both, to a multimodal large model, alongside masked semantic frames: The arguments that are highlighted with \textcolor{red}{red boxes} in the output structure are not provided as part of the input and have to be predicted.}
    \label{fig:input_output_cloze}
\end{figure*}

\paragraph{Data Annotation} Our annotation process occurs in three stages. \textit{Stage~1:} The goal of the first stage is to identify the implicit entities from the context in the form of either video or text. To achieve this goal, we hired two PhD students who had linguistics knowledge. They were trained to extract both implicit and explicit information and to identify where information might be unstated. The annotators were given the full videos, as well as the recipe steps and their corresponding timestamps so that they did not have to watch the full video. For instance, for (i)~\quotes{add onions and tomatoes to the blender and blend them} and (ii)~\quotes{add spices and garlic to the blender}, the annotators turn (ii) into \quotes{add spices and garlic to \textit{onions, tomatoes}}. 

\textit{Stage~2:} Multi-step instructions featuring these implicit entities are automatically converted into semantic role labels using GPT-4o-Mini. Specifically, we manually annotate five examples that show how to convert multi-step instructions to our scheme of semantic frames, i.e., \{\action, \what, \where\}. 
We include them as in-context examples in a chain-of-thought prompt for automatic annotation.

\textit{Stage~3:} In the final stage, the automatically generated labels undergo manual correction by an annotator (again a PhD student), ensuring the implicit information is accurately specified in the arguments and that all cooking ingredients are included, but any tools are ignored, as per our annotation guidelines. The guidelines, along with the prompt, are provided in the supplementary material.
\paragraph{Dataset Statistics} 
Table~\ref{tab:data_stats} presents various statistics of the \procsrl~dataset. We observe that the average number of implicit entities is 6.29 for \what~and 5.21 for \where~across the 2.5K semantic frames in 231 videos. There are no empty instances for \what~semantic role, but 54\% of \where~semantic roles are empty.
\begin{table}[t]
    \centering
    \scalebox{0.7}{%
    \begin{tabular}{l|r}
        \hline
        Name & Value \\
        \hline
        Number of videos & 231 \\
        Average/max video duration & 125.33/588.2 \\
        Average/max steps per video & 7.47/14 \\
        Average/max SRLs per video & 11.02/24 \\
        \hline
        Unique \action & 158 \\
        Unique entities & 805 \\
        \hline
        Unique \what\ arguments & 726 \\
        Average entities per \what & 4.45 \\
        Average implicit entities per \what & 6.29 \\
        \what~emptyset/total count & 0/2545 \\
        \hline
        Unique \where\ arguments & 626 \\
        Average entities per \where & 3.82 \\
        Average implicit entities per \where & 5.21 \\
        \where~emptyset/total count & 1393/2545 \\
        \hline
    \end{tabular}%
    }
    \caption{Statistics of \textbf{\procsrl}~ Dataset. The unit for video duration is \textit{seconds}.}
    \vspace{-1em}
    \label{tab:data_stats}
\end{table}
\section{Task}
We define two tasks on our \procsrl~Dataset, namely Semantic Argument Prediction and Next Step Prediction. Their goal is to use context to predict the semantic frames of instruction steps, focusing in particular on implicit arguments. For both tasks, we consider two different scenarios: the input is either (a) a sequence of textual instruction steps or (b) a sequence of trimmed video clips (one clip per step). We refer to this input as context~$\mathcal{C}$.

\paragraph{Implicit Argument Prediction}  
A cloze task involves filling in missing words or phrases in a text to measure understanding and inference abilities. \citet{rubin1976effectiveness} demonstrated that human accuracy in cloze tasks increases with the amount of context available and is also affected by where that context is positioned. Therefore, this task is suitable for assessing the contextual reasoning skills of a model; in our case, the ability to predict the entities in cooking recipes that fill semantic roles. 

We are given an input sequence $\mathcal{C}$ and corresponding semantic frames ${\{(c_i, a_i)\}}_{i=1:K}$. Here $c_i$ refers the single instruction in the text or trimmed video clip, $a_i = \{(v_l, wh_l, ww_l)\}_{l=1:L}$ is the annotated semantic frames. Here, $v_l$ refers to the \action-predicate, while $wh_l$ and $ww_l$ represent the arguments for \what~and~\where~sematic roles consisting of both explicit and implicit entities, i.e., $wh_l = \{e_{explicit} \cup e_{implicit}\}$. To evaluate these sets of arguments, we use the F1-score, for both the combined set and the set of implicit entities.

The cloze task is a sequence-to-sequence problem, where the input is the sequence $C$ and the corresponding semantic frames. We mask the arguments of the \what~and \where~semantic roles, while the predicate is given in the prompt, i.e., we have \srl{v_l}{[?]}{[?]}. The goal is to predict the masked arguments, which can include both explicit and implicit entities.
\paragraph{Next Step Prediction}
In this task, we follow the protocol of~\citet{Abdelsalam_2023_ICCV}: Given the partial input sequence $\mathcal{C}_{:t}$ for the first $t$ instruction steps either in a video or textual format, and the semantic role labels $(c_{1:t}, a_{1:t})$ for these steps, the task is to output $k$ plausible options for the next instruction step. The next step predictions are expressed as a natural language sentence in conjunction with the corresponding semantic role labels, i.e., $\{(s^{(1)}_{t+1}, a^{(1)}_{t+1}), \ldots , (s^{(k)}_{t+1}, a^{(k)}_{t+1})\}$. We generated 802 samples for next step prediction from 231 annotated instructional videos with $t\geq3$. 
\section{The \procsrlmodel~Model}
In this section we describe the~\procsrlmodel~model, which is designed to infer implicit argument information from the context in the form of multimodal inputs (video frames or text or both).
\paragraph{Silver-standard Dataset} To effectively infer implicit arguments from video, we create a silver-standard dataset. 
To achieve this, we augment the training dataset from Tasty~\cite{sener2022transferring} that already comes with temporally annotated instructions. By leveraging the chain-of-thought prompt along with the in-context examples we used for the annotation of the test set, we generate the silver-standard dataset for training a model that can identify implicit arguments. Specifically, we prompt the GPT-4o to perform two tasks similar to our manual annotation: (i)~split the multi-step instructions into single predicate-argument structures, (ii)~automatically infer the implicit entities and update the predicate arguments conditioned by the sequence of textual instructions. Finally, we generated $\sim$2.5K training instructional video samples and formatted the samples as required for next step prediction, leading to $\sim$18K training samples.
\paragraph{Training Objective}
We perform supervised instruction fine-tuning with LoRA of Qwen2-7B-Instruct and Qwen2-VL-7B-Instruct model, leveraging our auto-generated silver-standard dataset. In this work, we train a model with the next step prediction task, i.e. given a partial sequence of a cooking recipe (in text or video), the model is required to predict the next textual step. Additionally, it must generate semantic frame tuples that include both explicit and implicit entities.
\section{Experimental Setup}
\subsection{Implementation Details} 
\paragraph{Training and Inference} We perform fine-tuning with the default LoRA~\cite{hu2021lora} configurations provided in LLama-factory \cite{zheng2024llamafactory} on four A100-80GB GPUs with maximum duration of 48 GPU hours. Further implementation details are available in supplementary materials.
\subsection{Evaluation Metrics}
To assess the predictions of semantic arguments for \what\ and \where\ across both tasks, we employ the F1-score as the performance metric. Predicted and actual arguments are treated as sets of entities; the predicted entities are $P = \{w_1^p, w_2^p, \ldots, w_i^p\}$ and the gold-standard entities are $G = \{w_1^g, w_2^g, \ldots, w_j^g\}$, where each $w_i$ can be a phrase. We use the $\hat{\cap}(w_i^p, w_j^g)$ function (Equation~\ref{eq:iou_fn}) to compute precision ($\frac{|P\;\hat{\cap}\; G|}{|P|}$) and recall ($\frac{|P\;\hat{\cap}\; G|}{|G|}$).
We first identify all exact matches between the predicted and gold-standard entity sets and for non-exact matches, we calculate word overlap (intersection over union, IoU). We also compute the F1-score for implicit arguments separately.
\begin{align}%
\hat{\cap}(w_i^p, w_j^g) &= 
\begin{cases} 
1, & \text{if } w_i^p = w_j^g \\ 
\text{IoU}(w_i^p, w_j^g), & \text{otherwise}
\end{cases}\label{eq:iou_fn}
\end{align}%
\begin{table*}[t]
    \centering
    \scalebox{0.8}{%
    \renewcommand{\tabcolsep}{0.4cm}
    \begin{tabular}{c|l|c|c|c|c|c|c|c}
        \hline
        \multirow{2}{*}{\textbf{Row}} & \multirow{2}{*}{\textbf{Model}} & \multirow{2}{*}{\textbf{\#Params}} & \multirow{2}{*}{\textbf{Type}} & \multirow{2}{*}{\textbf{FT}} & \multicolumn{2}{|c|}{\textbf{SRL}} & \multicolumn{2}{|c}{\textbf{SRL-Implicit}} \\
        \cline{6-9}
        & & & & & $F1_{what}$ & $F1_{where}$ & $F1_{what}$ & $F1_{where}$ \\
        \hline
        \multicolumn{9}{|c|}{\textit{\textcolor{brown}{Proprietary Model}}} \\
        \hline
        1. & GPT-4o & - & T & \cross & 60.62 & 54.79 & 45.50 & 47.31 \\
        2. & GPT-4o & - & V & \cross & 49.63 & 42.02 & 43.38 & 40.08 \\
        3. & GPT-4o & - & V + T & \cross & 64.83 & 55.32 & 50.53 & 49.01 \\
        \hline
        \multicolumn{9}{|c|}{\textit{\textcolor{brown}{Small Open Source Models (<10B)}}} \\
        \hline
        % Random (Qwen2-VL) & 7B & - & 27.73 & 23.26 & 28.59 & 21.44 \\
        4. & Qwen2 & 7B & T & \cross & 44.51 & 26.94 & 19.88 & 25.81 \\
        5. & Qwen2-Math & 7B & T & \cross & 29.33 & 7.35 & 15.42 & 6.75 \\
        % Qwen2.5 & 7B & Text & 48.19 & 36.16 & 25.68 & 36.96 \\
        6. & LLama-3.1 & 8B & T & \cross & 52.05 & 38.27 & 31.22 & 39.11 \\
        \rowcolor{Cyan!10} %
        7. & \procsrlmodelT & 7B & T & \tick & 57.82 & 49.33 & 51.70 & 47.74 \\
        \hline
        8. & LongVA & 7B & V & \cross & 9.49 & 3.16 & 6.81 & 3.80 \\
        9. & LLava-OV & 7B & V & \cross & 20.78 & 7.86 & 11.43 & 8.44 \\
        10. & Qwen2-VL & 7B & V & \cross & 30.20 & 15.15 & 22.51 & 17.07 \\
        \rowcolor{Cyan!10} %
        11. & \procsrlmodel & 7B & V & \tick & 46.21 & 33.43 & 45.76 & 36.06 \\
        \hline
        12. & Qwen2-VL & 7B & V + T & \cross & 42.07 & 22.54 & 22.68 & 21.96 \\
        % Qwen2-VL & 7B & Video (Vision + LLM) & 42.11 & 31.08 & 39.77 & 34.50 \\
        \rowcolor{Cyan!10} %
        13. & \procsrlmodel & 7B & V + T & \tick & \textbf{64.86} & 54.54 & \textbf{59.15} & \textbf{56.21} \\
        \hline
        \multicolumn{8}{|c|}{\textit{\textcolor{brown}{Large Open Source Models (>10B)}}} \\
        \hline
        14. & Qwen2 & 72B & T & \cross & 56.10 & 50.86 & 41.82 & 49.61 \\
        15. & LLama-3.1 & 70B & T & \cross & 63.04 & \textbf{55.50} & 50.46 & 53.42 \\
        16. & LLava-OV & 72B & V & \cross & 39.52 & 29.05 & 34.34 & 36.48 \\
        17. & Qwen2-VL & 72B & V & \cross & 44.55 & 38.48 & 35.63 & 39.43 \\
        \hline
    \end{tabular}%
    }
    \caption{\textbf{Implicit Argument Prediction:} We report results for different input \textbf{Type}s: text-only (\textbf{T}), video-only (\textbf{V}), and video-text (\textbf{V+T}) for predicting \what~and \where~given the \action. \textbf{SRL} includes both explicit and implicit entities, while \textbf{SRL-Implicit} focuses only on implicit ones. \textbf{FT:} Fine-tuning with our silver-standard dataset. \textbf{\#Params:} Size of model.}
    \label{tab:reasoning}
\end{table*}
For the next step prediction task, we follow~\citet{abdelsalam2023gepsan} and predict $k$ plausible next step. Unlike the original method, we enhance the evaluation by employing a sliding window of the next three steps from the gold-standard instructions and use this to identify the best match based on the cosine similarity of embeddings. For the best match, we compute the verb recall $R_{verb}$@5 and the F1-score for semantic arguments as for the cloze task. Our task requires that semantic frames and instruction text are generated at the same time, so in addition to our SRL metrics, we also compute the standard natural language generation metrics BLEU4~\cite{papineni2002bleu} and METEOR~\cite{banerjee-lavie-2005-meteor} on text.
\subsection{Models}
\paragraph{Proprietary Model} We utilize the most recent OpenAI GPT-4o~\cite{hurst2024gpt} model (gpt-4o-2024-08-06); all the experiments were conducted before February 2025. We report the results on both text and vision versions of the model.
\paragraph{Open Source Models} We use decoder-only models for the evaluation of both tasks. For \textbf{text-only inputs}, we use Qwen2 instruct~\cite{yang2024qwen2} as the primary language model, due to its performance and use in recent video-LLMs. We also use the LLama 3.1~\cite{dubey2024llama} instruct model. For \textbf{video-only inputs}, we focus on models with long context, i.e., models accepting a large number of frames. Specifically, we utilize LongVA~\cite{zhang2024long} and Qwen2-VL~\cite{Qwen2VL}. Given Qwen2-VL's capability to handle a context of 32K tokens, we designate it as the primary model for both tasks. In addition to this, we also evaluate LLava-OneVision \cite{li2024llava} using recursive inference. This involves predicting semantic arguments for a single instruction based on a given input modality and then prepending the response to the prompt for the subsequent prediction. For video inputs, we limit to 320 frames per video at the highest resolution based on available GPU memory.
\section{Results and Discussion}
\subsection{Implicit Argument Prediction}
Our experimental results for semantic argument prediction, including implicit entities for \what~and \where~semantic roles, are summarized in Table~\ref{tab:reasoning}. The results are organized based on the model types: Proprietary Model (Rows~1--3), Small Open Source Models (Rows~4--13), and Large Open Source Models (Rows~14--17). 

\begin{figure*}[t]
    \centering
    % \captionsetup{font=footnotesize}
    \includegraphics[width=0.9\linewidth]{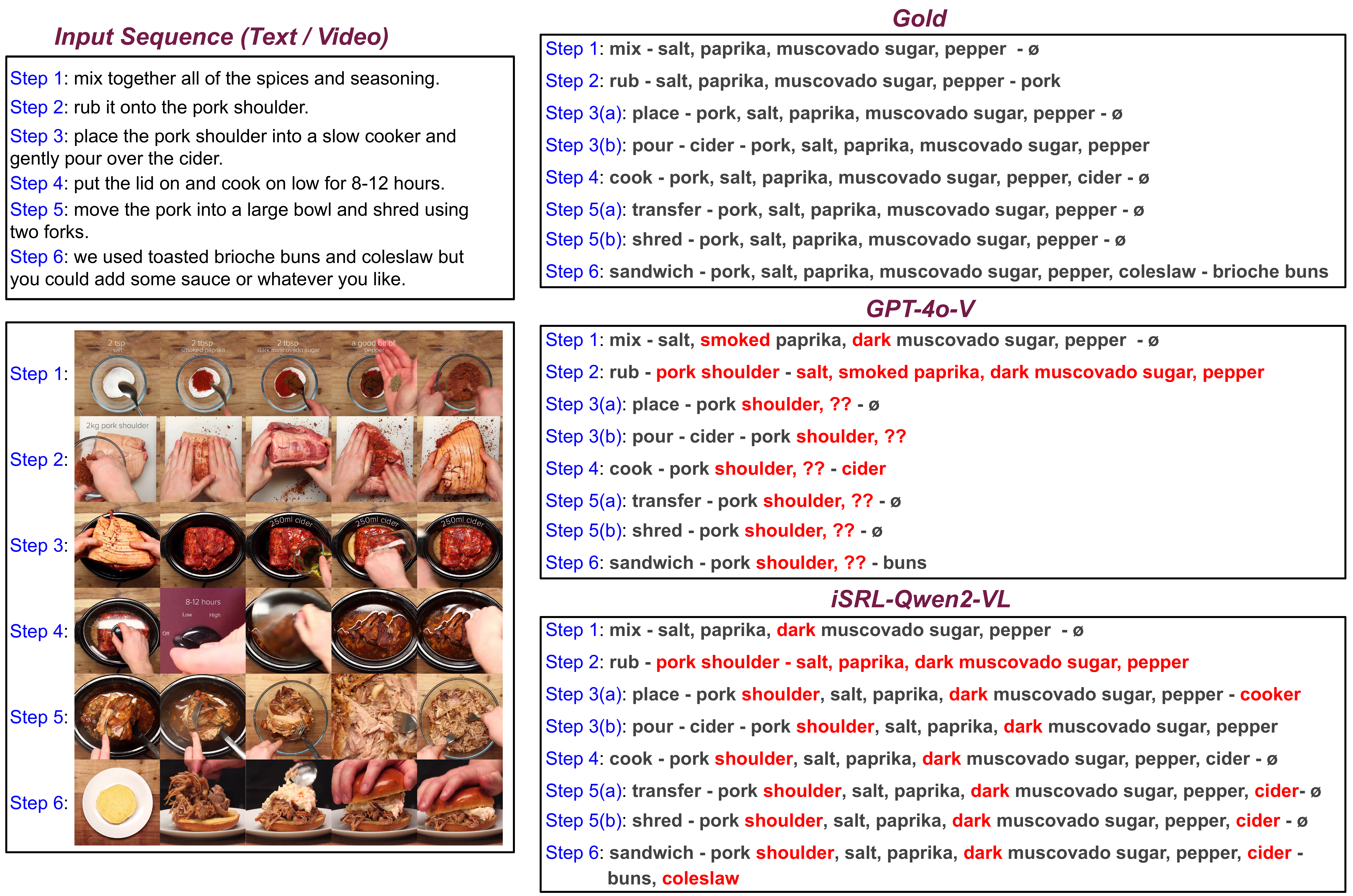}
    \caption{\textbf{Qualitative example using video-only predictions.} The example is from TASTY~\cite{sener2022transferring} with ID-\href{https://tasty.co/recipe/cider-pulled-pork}{cider-pulled-pork}. The examples highlight common errors in the predictions, i.e., a failure to track the mixture ingredients, as in step~3(a) pork is mixed with spices. Incorrect predictions are highlighted in \textcolor{red}{red} and missing ingredients are indicated using \textcolor{red}{`??'.}}
    \label{fig:example_result}
\end{figure*}

In the scenario with text-only input, we observe that LLama-3.1-70B (Row~15) achieves the highest performance for implicit argument prediction, closely followed by GPT-4o (Row~1). Notably, our Small \procsrlmodelT~model (Row~7) also shows competitive results in implicit argument prediction.

When evaluating the scenario with video-only input, large models such as GPT-4o (Row~2) and Qwen2-VL (Row~17) achieved the best F1-scores, although their performance is lacking behind the scenario with text-only input. This highlights two primary challenges in semantic argument prediction from video: recognizing entities in the current step and inferring or tracking entities within the temporal context. For smaller open-source models, performance is poor. However, our model~\procsrlmodel~bridges the gap to text-only models by effectively doubling the results when using video-only input, i.e., $F1_{\iwhat}$ and $F1_{\iwhere}$ improve by 23.25\% and 18.99\% (comparing Rows~10 and 11).
\begin{table*}[t]
    \centering
    \scalebox{0.8}{%
    \renewcommand{\tabcolsep}{0.3cm}
    \begin{tabular}{|c|l|c|c|c|c|c|c|c|c|c}
        \hline
        \multirow{2}{*}{\textbf{Row}} & \multirow{2}{*}{\textbf{Model}} & \multirow{2}{*}{\textbf{\#Params}} & \multirow{2}{*}{\textbf{Type}}& \multirow{2}{*}{\textbf{FT}} & \multicolumn{3}{|c|}{\textbf{SRL Prediction}} & \multicolumn{2}{|c|}{\textbf{Sentence Predictions}} \\
        \cline{6-10}
        & & & & & $R_{verb}$@5 & $F1_{what}$ & $F1_{where}$ & B4 & METEOR \\
        \hline
        1. & GEPSAN & - & T & - & - & - & - & 0.30 & 9.88 \\
        % GepSan & - & Video & - & - & - & - & - \\
        2. & GPT-4o & - & T & \cross & \textbf{52.05} & 18.48 & 15.04 & 3.18 & 17.93 \\
        3. & Qwen2 & 72B & T & \cross & 47.84 & 17.59 & 13.44 & 4.92 & 20.22 \\
        4. & Qwen2 & 7B & T & \cross & 45.39 & 9.49 & 9.63 & 3.34 & 17.86 \\
        \rowcolor{Cyan!10} %
        5. & \procsrlmodelT & 7B & T & \tick & 50.01 & \textbf{20.29} & \textbf{15.99} & \textbf{6.48} & \textbf{20.54} \\
        \hline
        6. & GPT-4o & - & V & \cross & \textbf{49.89} & \textbf{18.02} & 16.14 & 2.14 & 16.20 \\
        7. & Qwen2-VL & 72B & V & \cross & 41.19 & 14.76 & 12.42 & 1.56 & 15.31 \\
        8. & Qwen2-VL & 7B & V & \cross & 40.41 & 8.33 & 9.47 & 2.05 & 15.77 \\
        \rowcolor{Cyan!10} %
        9. & \procsrlmodel & 7B & V & \tick & 44.56 & 16.22 & \textbf{16.58} & \textbf{3.69}& \textbf{17.63} \\
        \hline
        10. & GPT-4o & - & V + T & \cross & \textbf{53.36} & \textbf{20.51} & 16.32 & 4.34 & 18.99 \\
        11. & Qwen2-VL & 7B & V + T & \cross & 38.56 & 7.62 & 9.43 & 2.40 & 16.18 \\
        \rowcolor{Cyan!10} %
        12. & \procsrlmodel & 7B & V + T & \tick & 47.76 & 19.74 & \textbf{17.44} & \textbf{5.22} & \textbf{19.38} \\
        \hline
    \end{tabular}%
    }
    \caption{\textbf{Next Step Anticipation.} We report results for input \textbf{Types}: text-only (\textbf{T}), video-only (\textbf{V}), and video-text (\textbf{V+T}) to predict \{\action,\what,\where\} with explicit and implicit arguments, plus natural text for the next step. \textbf{SRL prediction} evaluates the predicate-argument structure, while \textbf{Sentence Predictions} uses NLG metrics for predicted text. \textbf{FT:} Fine-tuning with our silver-standard dataset. \textbf{\#Params:} Size of model.}
    \label{tab:next_step}
\end{table*}

We examine different models with multimodal input (V+T) and found that it enhances performance over single modality inputs to infer implicit information. Presumably, multimodal input makes it possible to recognize local entities and to resolve visual-linguistic ambiguities, as shown in Figure~\ref{fig:teaser}. However, with both modalities, GPT-4o exhibits a bias towards textual entities, incorrectly specifying \textit{dressing} as an argument for \what, instead of the mixture's ingredients (see Figure~\ref{fig:example2_result}). In contrast, our model (Row~13), using multimodal input, achieves the best performance on the implicit metrics, effectively tracking and inferring implicit entities from the temporal context. Overall, \procsrlmodel\ achieves a 17\% relative improvement in F1-score for \iwhat~and a 14.7\% for \iwhere~semantic roles over GPT-4o.
\paragraph{Qualitative Results} The example in Figure~\ref{fig:example_result}, highlights common errors in predictions. It includes instructional text sequences and video frames, comparing video-only predictions from GPT-4o and our \procsrlmodel~model with gold-standard semantic frames. A frequent error is the models' inability to temporally track ingredients. For instance, in step~3(a), the \what~arguments are spices rubbed on pork, but the model overlooks that this pork is mixed with spices, differing from step~2. Such inferences are crucial for understanding long temporal contexts due to visual dynamics. In contrast, \procsrlmodel~effectively tracks ingredients but misses that pork with spices is removed from the cider before shredding it in step~5.
\subsection{Next Step Prediction}
The results for the next step prediction task are shown in Table~\ref{tab:next_step}. Sentence prediction metrics emphasize text fluency without considering implicit information, whereas SRL predictions assess the model's ability to infer implicit arguments. The most important observation is that adding our \emph{iSRL} approach to the corresponding base model consistently and significantly improves performance across all metrics (Row 4/5, 8/9, and 11/12). Overall, we achieve state-of-the-art performance across modalities on NLG metrics and remain competitive with, or better than, large models in semantic argument predictions. The text-only and multimodal models (Qwen2/GPT-4o) perform better than the specialized model (Row~1) for next step prediction trained on Recipe1M+~\cite{marin2021recipe1m+} dataset. 
\subsection{Ablation and Analysis}
\paragraph{Chain-of-thought Prompt} We experimented with various prompts to enhance the performance of the Qwen2 model using both text-only and video input, as detailed in Table~\ref{tab:srl_cot}. The chain-of-thought prompt defines semantic roles alongside implicit and explicit entities, summarizing our annotation scheme as a thought sequence. Table~\ref{tab:srl_cot} shows that chain-of-thought (CoT) prompting significantly improves \where~argument identification.
Without our CoT prompt, the model tends to focus on cooking tools for the \where~semantic role. This highlights the effectiveness of our structured prompt in guiding the model to a more accurate understanding of the task.
\paragraph{Fine-tuning Input} We investigated the impact of fine-tuning the Qwen2 model with text-only input for next step prediction task, with and without semantic role prediction. The results in Table~\ref{tab:srl} reveal that naive finetuning without incorporating SRL predictions in the output diminishes the model's ability to infer and track explicit and implicit arguments. Conversely, when SRL predictions are included, the model's performance significantly improves, nearly doubling in effectiveness. Notably, for the $F1_{\iwhat}$ metric, the performance sees an impressive increase of 31.82\% points. 
\begin{table}[h!]
    \centering
    \scalebox{0.7}{%
    \begin{tabular}{c|c|c|c|c|c}
        \hline
        \multirow{2}{*}{Model} & \multirow{2}{*}{Pompt} & \multicolumn{2}{|c|}{SRL} & \multicolumn{2}{|c}{SRL-Implicit} \\
        \cline{3-6}
            & & $F1_{what}$ & $F1_{where}$ & $F1_{what}$ & $F1_{where}$ \\
        \hline
        \multirow{2}{*}{Qwen2} & CoT & 44.51 & 26.94 & 19.88 & 25.81 \\
        & w/o CoT & 42.90 & 20.60 & 18.02 & 19.30 \\
        \hline
        \multirow{2}{*}{Qwen2-VL} & CoT & 30.20 & 15.15 & 22.51 & 17.07 \\
        & w/o CoT & 30.09 & 8.60 & 22.93 & 9.95 \\
        \hline
    \end{tabular}%
    }
    \caption{Impact of Chain-of-Thought Prompting (CoT).}
    \label{tab:srl_cot}
\end{table}
\begin{table}[h!]
    \centering
    \scalebox{0.7}{%
    \begin{tabular}{c|c|c|c|c|c}
        \hline
        \multirow{2}{*}{Model} & \multirow{2}{*}{FT} & \multicolumn{2}{|c|}{SRL} & \multicolumn{2}{|c}{SRL-Implicit} \\
        \cline{3-6}
            & & $F1_{what}$ & $F1_{where}$ & $F1_{what}$ & $F1_{where}$ \\
        \hline
        \multirow{3}{*}{Qwen2} & Zero-shot & 44.51 & 26.94 & 19.88 & 25.81 \\
        & w/o SRL & 21.90 & 13.17 & 8.38 & 10.88 \\
        & w/ SRL & 57.82 & 49.33 & 51.70 & 47.74 \\
        \hline
    \end{tabular}%
    }
    \caption{Effectiveness of fine tuning (\textbf{FT}) with \textbf{SRL}. \textbf{Zero-shot} is the performance without fine tuning.}
    \label{tab:srl}
\end{table}
\begin{figure}[h!]
    \centering
    \scalebox{0.5}{%
    \begin{subfigure}[b]{0.45\textwidth}
        \includegraphics[width=\linewidth]{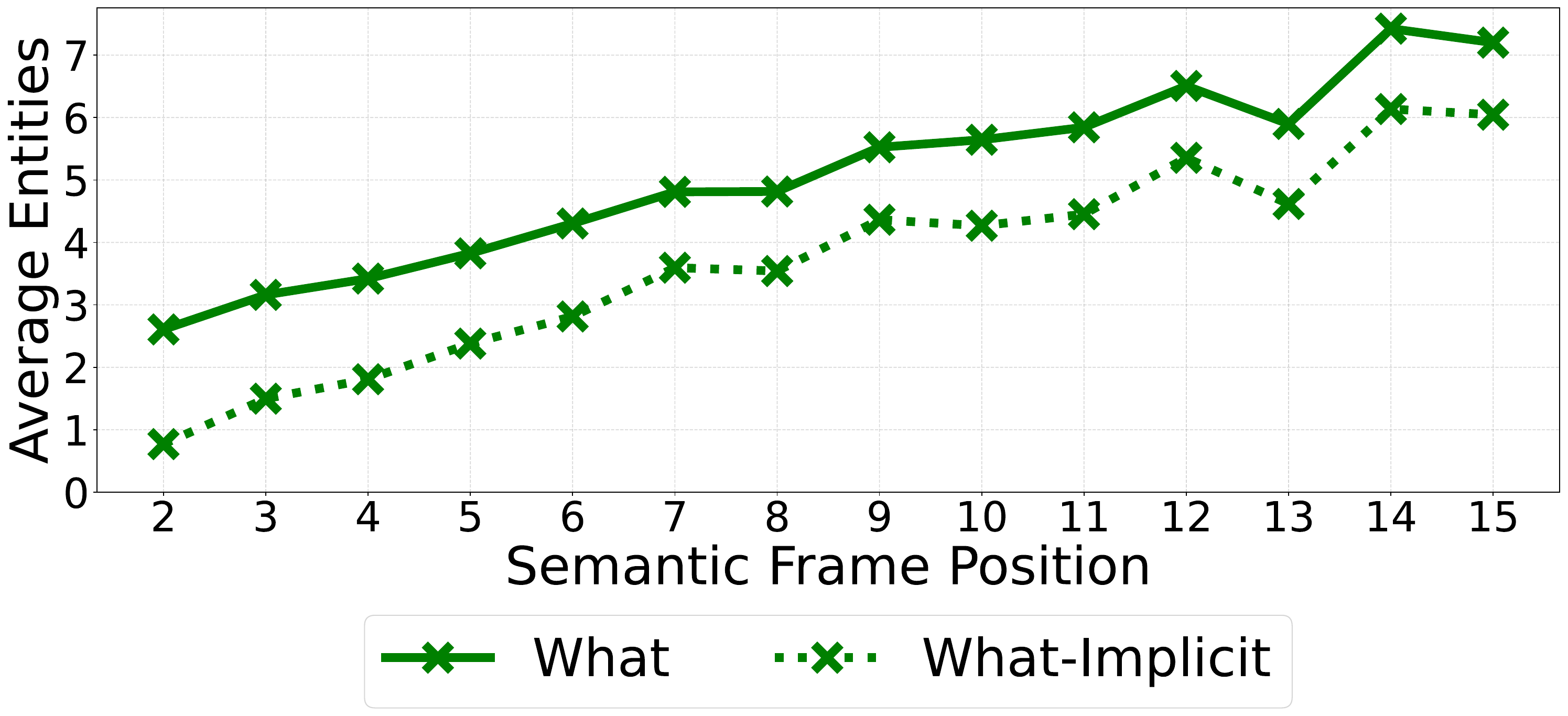}
        \caption{}
        \label{fig:length_analysis_a}
    \end{subfigure}
    \hfill
    \begin{subfigure}[b]{0.45\textwidth}
        \includegraphics[width=\linewidth]{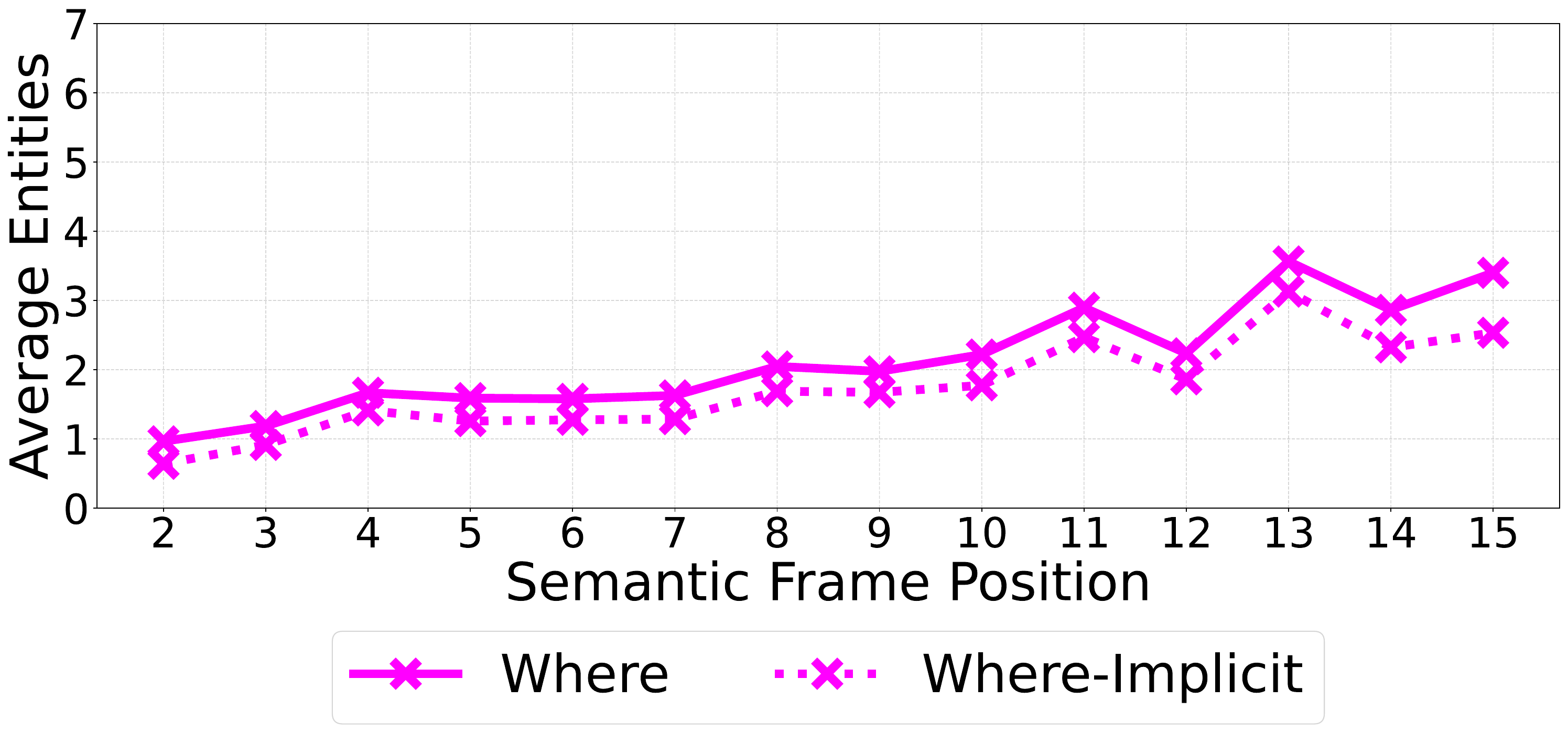}
        \caption{}
        \label{fig:length_analysis_b}
    \end{subfigure}
    }

    \scalebox{0.5}{%
    \begin{subfigure}[b]{0.45\textwidth}
        \includegraphics[width=\linewidth]{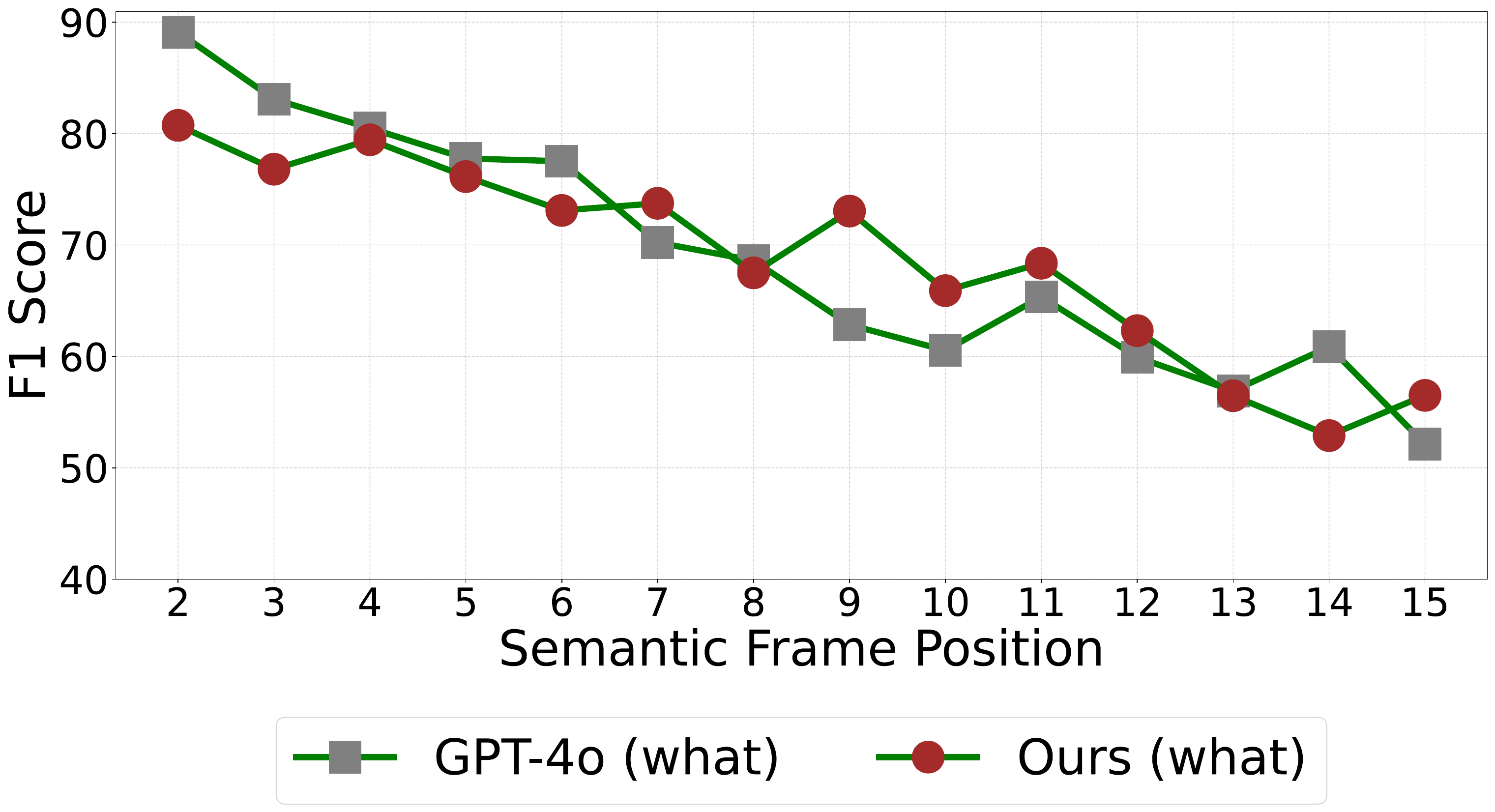}
        \caption{}
        \label{fig:length_analysis_c}
    \end{subfigure}
    \hfill
    \begin{subfigure}[b]{0.45\textwidth}
        \includegraphics[width=\linewidth]{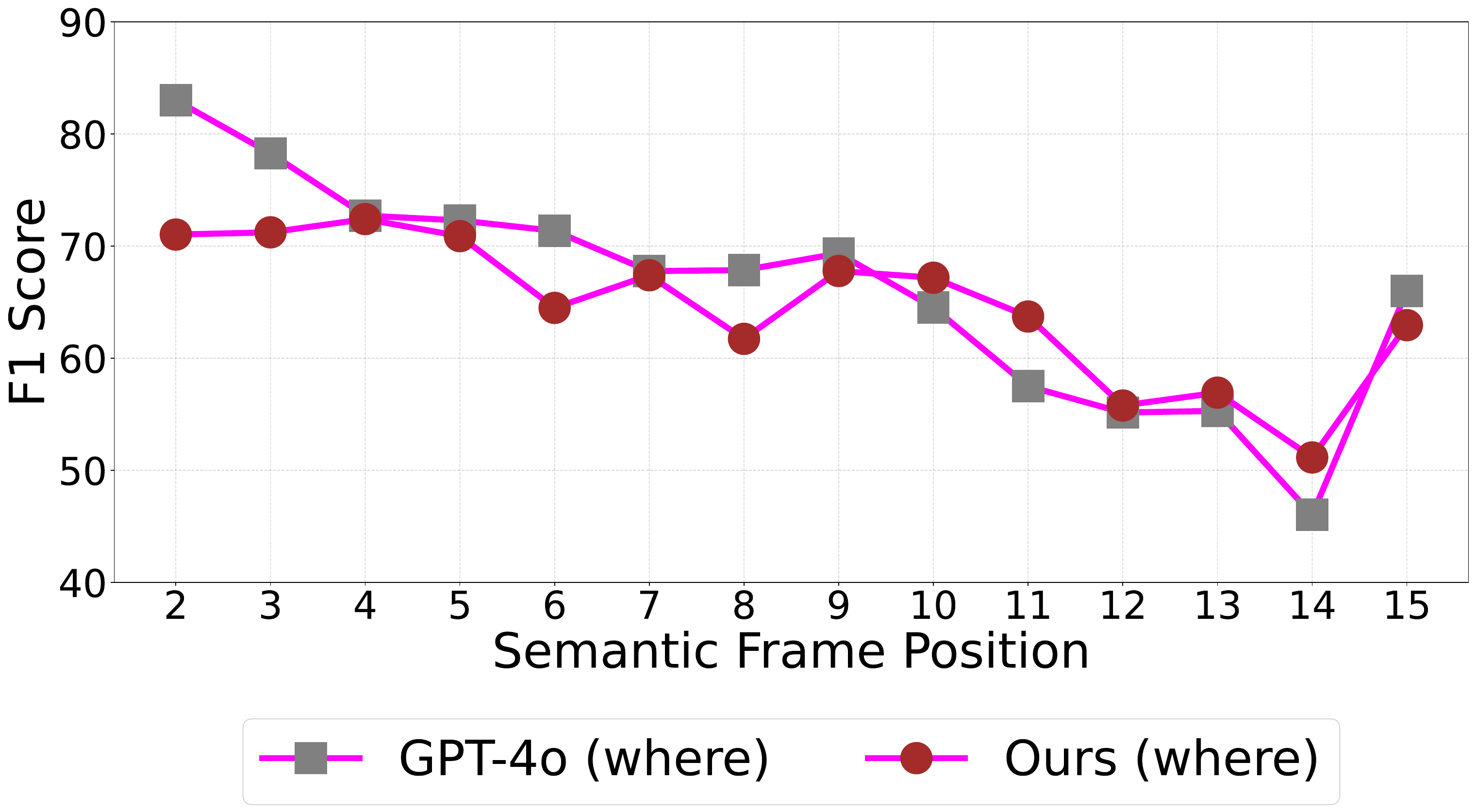}
        \caption{}
        \label{fig:length_analysis_d}
    \end{subfigure}
    }

    \scalebox{0.5}{%
    \begin{subfigure}[b]{0.45\textwidth}
        \includegraphics[width=\linewidth]{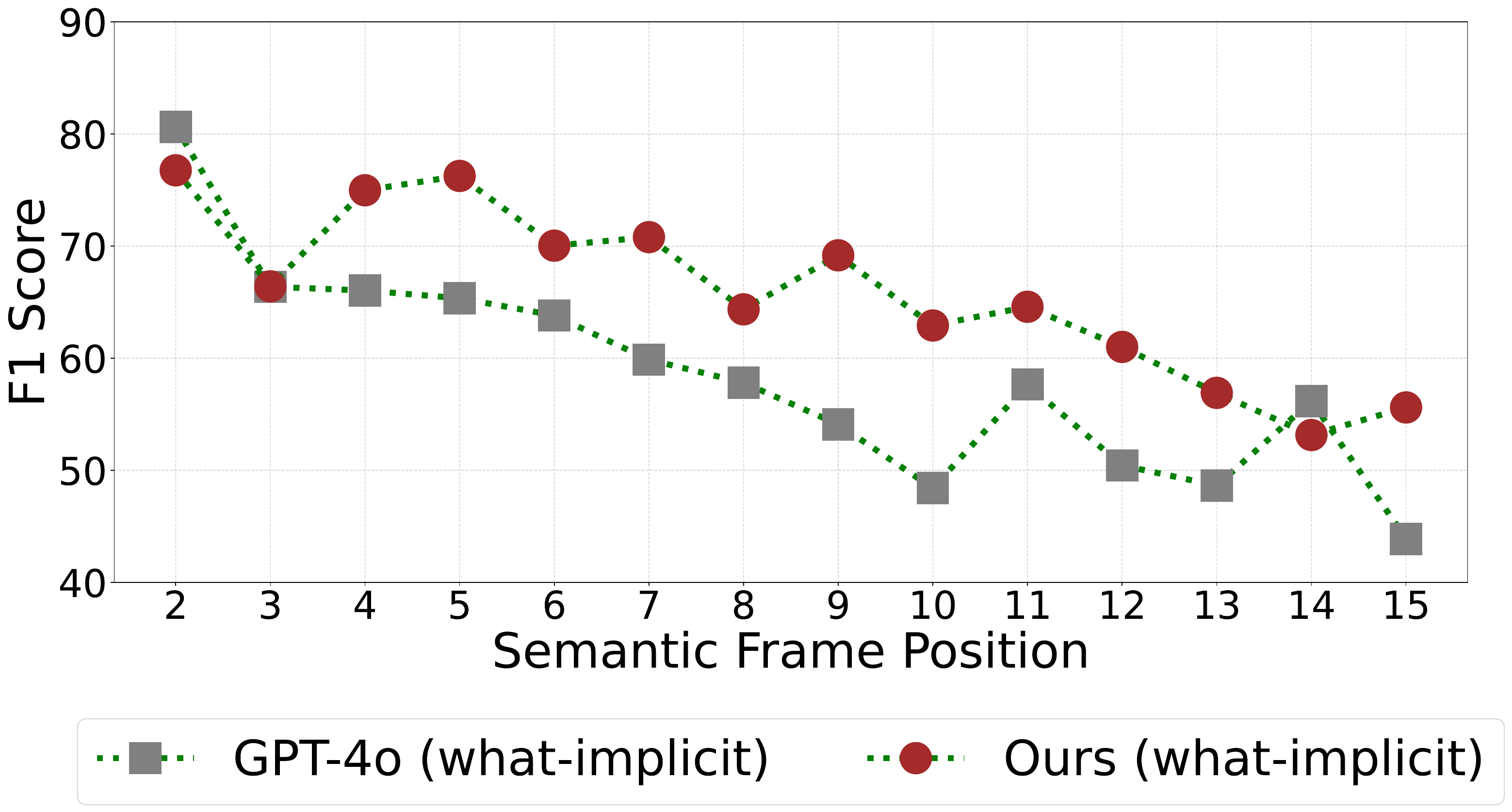}
        \caption{}
        \label{fig:length_analysis_e}
    \end{subfigure}
    \hfill
    \begin{subfigure}[b]{0.45\textwidth}
        \includegraphics[width=\linewidth]{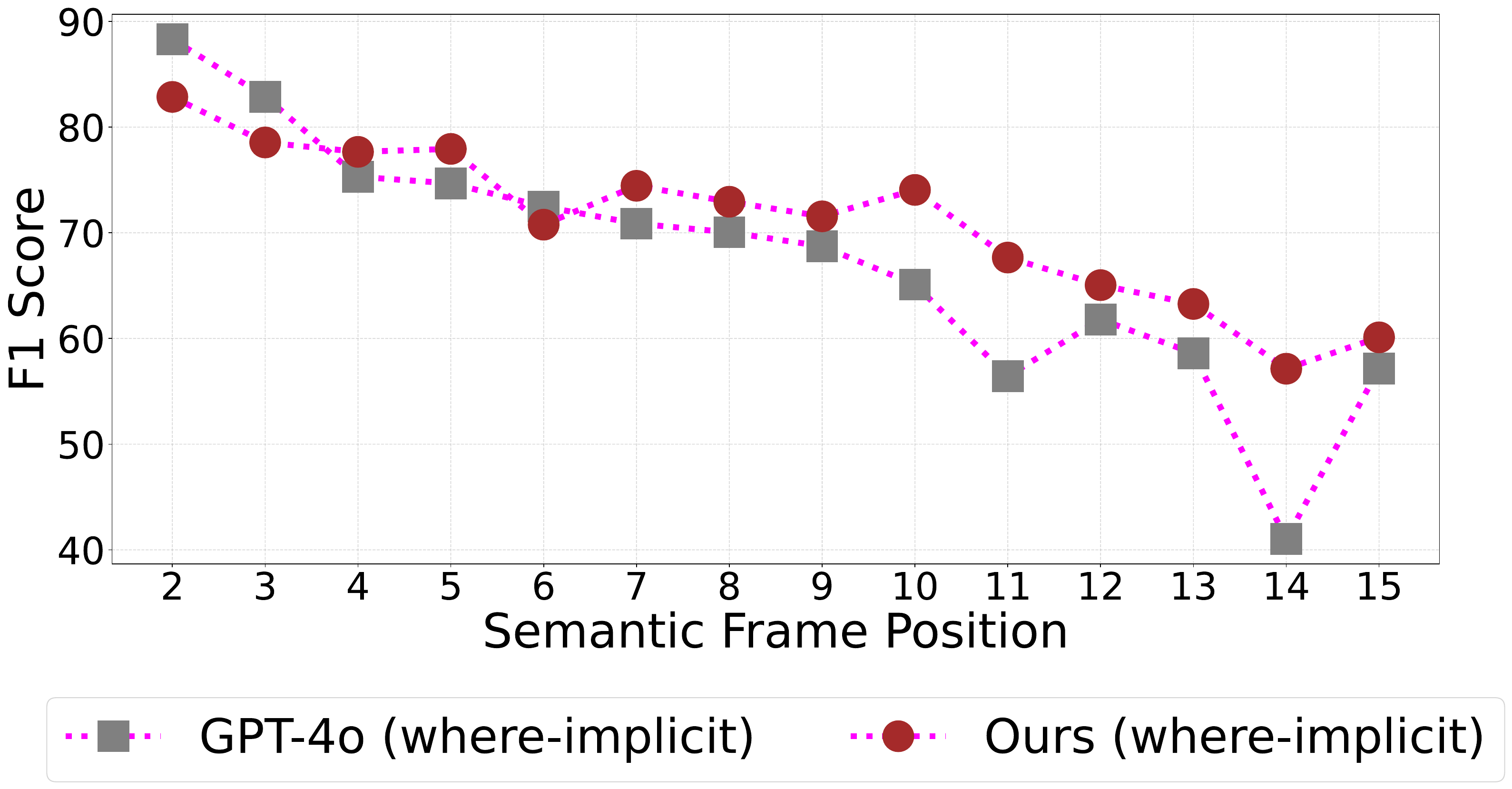}
        \caption{}
        \label{fig:length_analysis_f}
    \end{subfigure}
    }
    \caption{Comparison of GPT-4o and our~\procsrlmodel~model for argument prediction across \textit{semantic frame} positions in multi-modal procedural inputs (V+T).}
    \label{fig:length_analysis}
\end{figure}
\paragraph{Effect of Semantic Frame Position} Figure~\ref{fig:length_analysis_a},\ref{fig:length_analysis_b} illustrates that the average number of entities overall, as well as the average number of implicit arguments. Both averages are higher in later semantic frame positions, requiring models to use temporal context for accurate tracking and prediction of semantic arguments. This emphasizes that tracking entities in our proposed dataset, \procsrl, becomes challenging as sequence length increases.

We also study the effect of argument prediction across the semantic frame position in multimodal procedural inputs (\textit{video and text}) for two models, i.e., GPT-4o and ours~\procsrlmodel. Figure~\ref{fig:length_analysis_c},~\ref{fig:length_analysis_d} presents \textit{SRL} prediction score and shows that both GPT-4o and \procsrlmodel~achieve strong results at early semantic positions, but performance decreases gradually as semantic position increases. In contrast, \procsrlmodel~achieves performance for implicit arguments across the board for \textit{what} arguments, and for later semantic frame positions for \textit{where} arguments (see Figure~\ref{fig:length_analysis_e},~\ref{fig:length_analysis_f}). This shows that while GPT-4o excels at local contextual reasoning, \procsrlmodel~is robust for longer contextual reasoning and better at tracking implicit entities. Refer to the appendix (see Section~\ref{sec:sup_results}) for video-only inputs.
\section{Conclusion}

In this work, we introduced \procsrl, a semantically annotated dataset for procedural video understanding, focusing on inferring implicit information from multimodal contexts. Our dataset is a step towards interpreting elliptical instructions, facilitating applications like personalizing cooking instruction (e.g.,~to account for allergies) and tracking entities in human-robot interaction. Through a cloze task, we assessed the ability of recent multimodal models, including GPT-4o, to infer implicit information and evaluate contextual reasoning using our dataset. Our experiments reveal that these models have difficulty tracking implicit entities. We used a chain-of-thought prompting approach with in-context examples to automatically generate a silver-standard dataset of semantic frames. We then proposed \procsrlmodel, a model fine-tuned with our silver-standard data, to predict semantic frames as intermediate representations. We tested the model for next step prediction and showed that it enhances contextual reasoning and entity tracking ability in longer sequences.

\section*{Limitations}
While our work advances multimodal procedural understanding, we discuss a few limitations. Automated semantic role labeling (SRL) uses GPT-4o and may introduce biases or inaccuracies in the SRL distribution.
The proposed method relies heavily on the quality of the initial SRLs --- suboptimal SRLs can lead to inaccurate implicit argument predictions and contextual reasoning skills.
Our method also assumes that the multi-step instructions of cooking recipes can be meaningfully decomposed into simple predicate-argument structure, which may not hold for other procedural data and require further decompositions.

\section*{Acknowledgements} This work was supported in part by the UKRI Centre for Doctoral Training in Natural Language Processing, funded by UKRI grant EP/S022481/1 and the University of Edinburgh, School of Informatics. Marcus Rohrbach was funded in part by an Alexander von Humboldt Professorship in Multimodal Reliable AI sponsored by Germany's Federal Ministry for Education and Research. 

% Bibliography entries for the entire Anthology, followed by custom entries
%\bibliography{anthology,custom}
% Custom bibliography entries only
\bibliography{custom}

\appendix
\clearpage
\section*{Appendix}
The supplementary material contains the following:
\begin{enumerate}
    % \item Additional Details of the~\procsrl dataset (Section~\ref{sec:sup_data});
    \item Annotation Guidelines (Section~\ref{sec:sup_guidelines});
    \item Additional Dataset sources and Implementation details (Section~\ref{sec:sup_impl}).
    \item Prompt Details for Silver-standard dataset and Task inference (Section~\ref{sec:sup_prompts});
    \item Additional Analysis and Qualitative Results (Section~\ref{sec:sup_results});
\end{enumerate}

\section{The \procsrl~ Dataset - Annotations}\label{sec:sup_guidelines}

\subsection{Annotation Guidelines}
\begin{figure*}[t]
    \centering
    \includegraphics[width=0.99\linewidth]{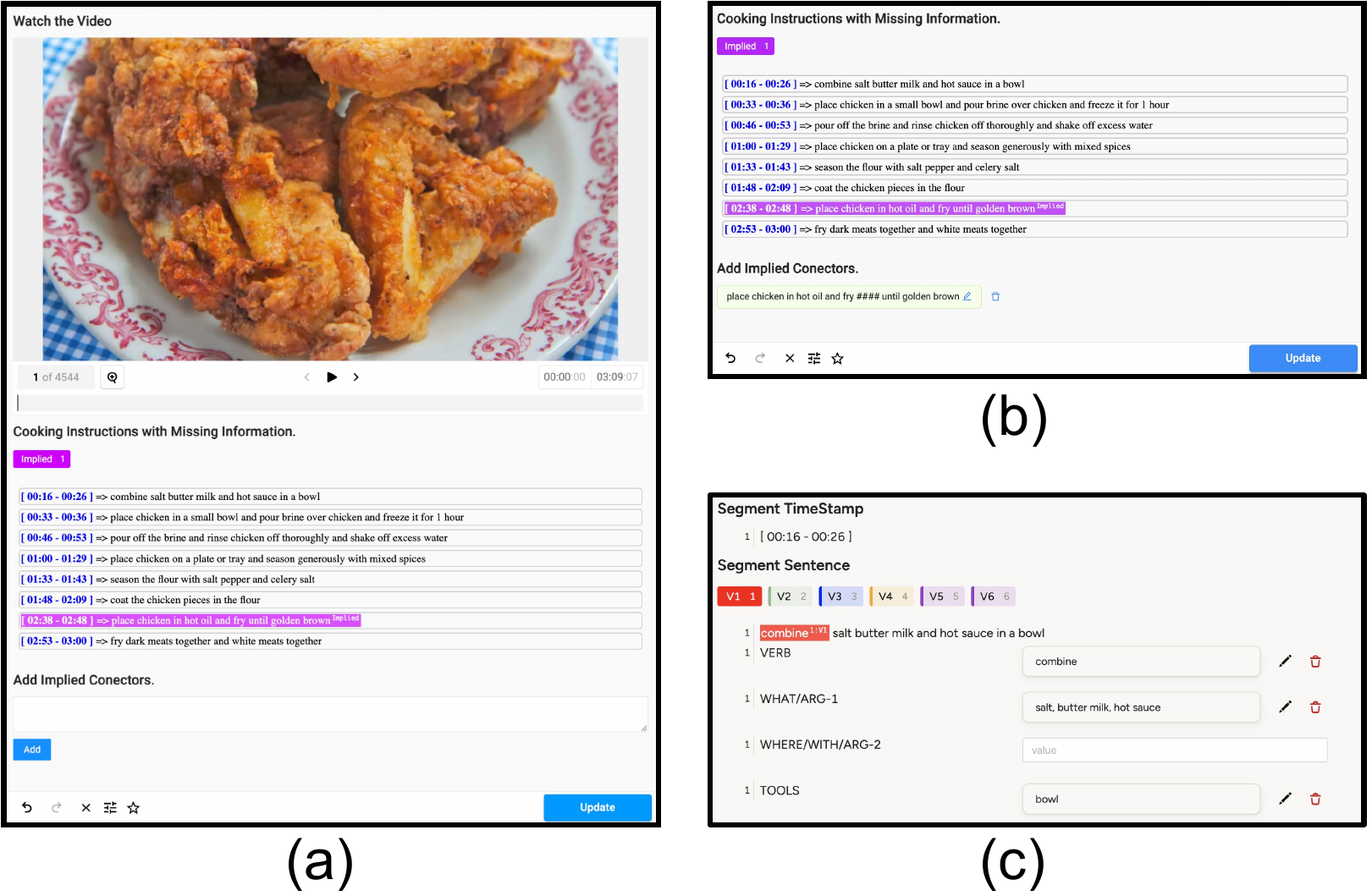}
    \caption{\textbf{Annotation Tool}. The images in (a) \& (b) shows the tool interface to annotate the implicit entities during Stage~1. While the image in (c) shows the tool interface for semantic role labeling in Stage~3 (the video is omitted for clarity).}
    \label{fig:tool}
\end{figure*}

\begin{figure*}[t]
    \centering
    % \captionsetup{font=footnotesize}
    \includegraphics[width=0.99\linewidth]{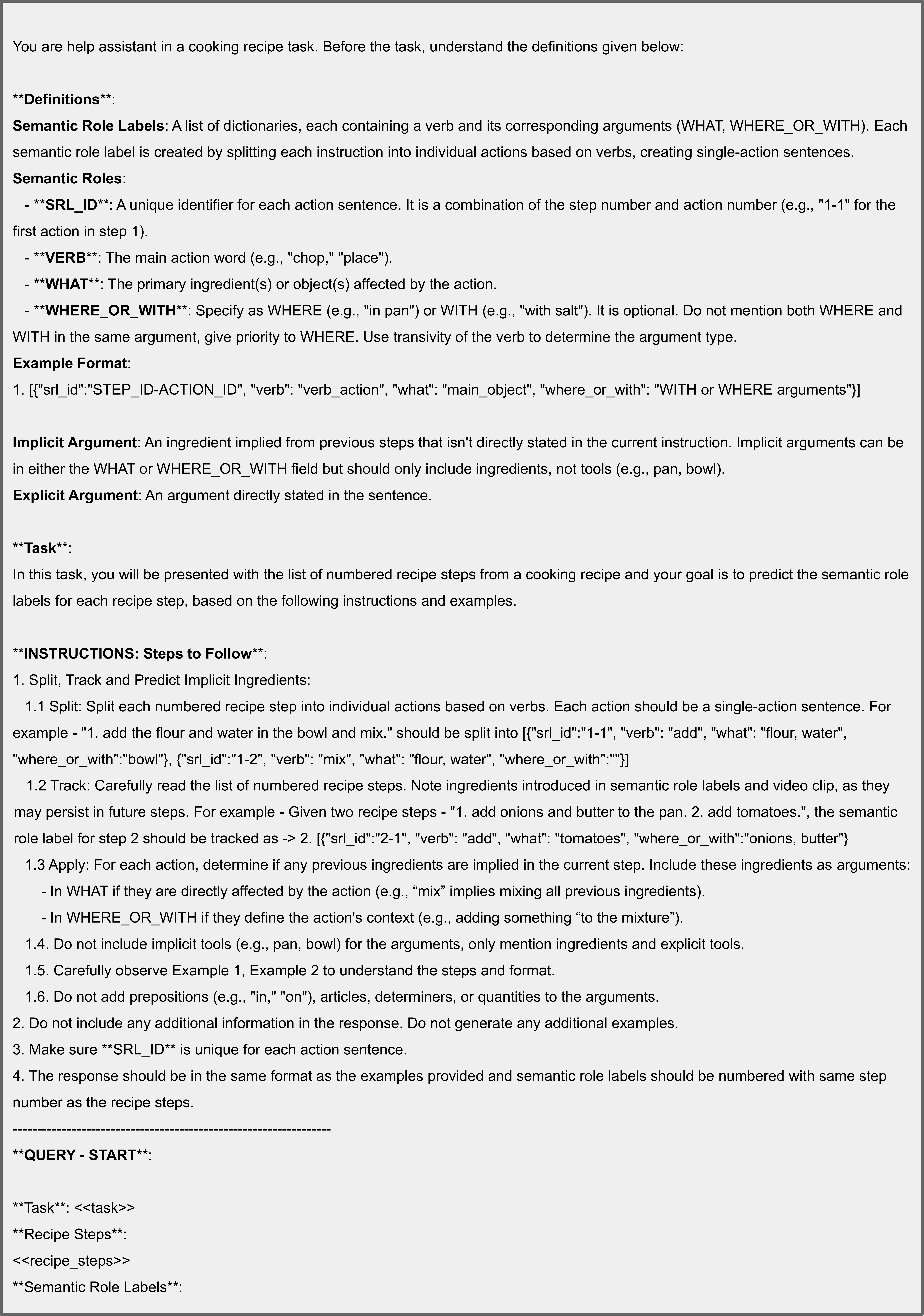}
    \caption{Dataset Prompt.}
    \label{fig:dataset_prompt}
\end{figure*}

The annotation tool is shown in Figure~\ref{fig:tool}
\paragraph{Stage 1 - Implicit Entities:} We define the Implicit Argument as an ingredient implied by earlier steps or video clips, not directly visible in the current clip or text, but it must appear in either the text or video of a previous step. The Explicit Argument is an argument directly mentioned or visible in the current video clip. The additional annotation guidelines are as follows.
\begin{enumerate}
    \item Focus only on cooking ingredients and ignore the tools or cooking utensils.
    \item Only use the previously mentioned entities for the implicit arguments.
    \item Only add implicit entities for nouns and pronouns like \quotes{them} and \quotes{it}, and disregard all other cases.
    \item If the verb is di-transitive i.e. need two objects such as \{\action,\what,\where\}. For example: in the sentence \quotes{add spices and onions to the bowl}, the verb \quotes{add} has two arguments \quotes{spices and onions} (i.e. items need to be added) and \quotes{the bowl} (i.e. where the items will be added. However in the sentences \quotes{add spices and onions} the second argument is missing, which is implicit and we need to add the implicit argument.
    \item If the verb is transitive i.e. need one object. For example, (i) \quotes{add flour and spices in a bowl and mix}. Here \quotes{mix} verb requires an implicit object i.e. mix flour and spices.
\end{enumerate}
\paragraph{Stage 2 - GPT-4o-Mini Labels:} The prompt to pre-annotate and split the multi-step instructions is shown in Figure~\ref{fig:dataset_prompt}.
\paragraph{Stage 3 - Manual Refinement:} Along with the Stage 1 guidelines, we share the following additional instructions for semantic role label refinement.
\begin{enumerate}
    \item Use previously mentioned entities for implicit arguments. For instance, (i) \quotes{add onions and tomatoes to the blender and blend them.}, (ii) \quotes{add spices and garlic to the blender} the second instruction becomes \srl{add}{[spices, garlic]}{[onions, tomatoes]}, as the blender already contains onions and tomatoes.
    \item Ignore instructions without \what~and \where~arguments, such as ``repeat''.
    \item Avoid instructions without visible ingredients like "turn heat to low" unless they are visible in the video. However, instructions like "bake tomatoes" are considered even if they are implicit.
    \item Determine \what~based on the verb. For example, \quotes{add sausages on top of potatoes} becomes \srl{add}{sausages}{potatoes}, and \quotes{top potatoes with sausages} turns into \srl{top}{potatoes}{sausages}.
\end{enumerate} 

\subsection{Annotator Recruitment and Payments}
We hired three final-year PhD students to annotate the implicit entities in Stage 1. Initially, we conducted a session to explain the task, guidelines, and provide a walk-through of the Label Studio~\footnote{https://labelstud.io/} tool. Despite this, one annotator struggled due to a lack of linguistic expertise. To address this, we refined our guidelines and held interactive sessions to resolve any queries. We equally divide the 700 samples among two annotators, and there is no overlap, i.e., each annotator worked on 350 samples each. This allowed us to continue effectively with the two annotators who had strong linguistic skills. They completed the work within 55 hours, and the entire process cost approximately \$1200. In the final stage, the automatically generated semantic frames using  GPT-4o-Mini are verified and corrected by an annotator who is different from the annotators involved in the Stage 1. The PhD student spent 40 hours reviewing and refining the semantic role labels generated after Stage 2, costing approximately \$800. This multi-stage annotation scheme ensures that the semantic frames are accurate. All annotators were trained in-house annotators (PhD students).
\section{Source Datasets and Implementation Details}\label{sec:sup_impl}
\paragraph{Source Datasets:} We utilize two instructional video datasets, YouCook2~\cite{ZhXuCoAAAI18} and Tasty~\cite{sener2022transferring}. These datasets consist of cooking videos with detailed annotations specifying instruction steps in English sentences along with temporal boundaries. YouCook2 videos are characterized by an unconstrained environment and a third-person viewpoint, while Tasty videos are captured using an overhead camera. The videos in YouCook2 are sourced from YouTube and there are on average $\approx$7.7 instructions per video. The videos in Tasty are relatively short, with an average duration of 54 seconds, and each recipe incorporates an average of nine instructions. An illustrative example can be found on the Tasty website\footnote{https://tasty.co/}.

Initially, we selected videos from YouCook2's validation set and Tasty's test set based on the following criteria: video duration (between 30 seconds and 10 minutes), at least 4 instructions, and valid YouTube video. Later, we selected 165 videos from Tasty's test set and 66 from YouCook2, focusing on those containing implicit information obtained after stage 1 of annotation. 

The Tasty dataset is a collection of 2511 unique recipes distributed across 185 tasks, such as making cakes, pies and soups, which is used to fine tune our \procsrlmodel~model after generating silver-standard semantic labels. 

\paragraph{Implementation Details:} We fine-tune our model with the default configurations provided in LLama-factory \cite{zheng2024llamafactory} specifically for LoRA~\cite{hu2021lora}. We perform distributed training on four A100-80GB GPUs with maximum duration of 48 GPU hours. This involves setting the rank to 8, a learning rate of $10^{-4}$, and a batch size of 64. For Qwen2-VL, we use video inputs to predict the next step, with a batch size of 16 and a learning rate of $10^{-5}$. Eight frames per clip are sampled at 224 pixels. The text model is fine-tuned for three epochs and the video model for two, taking 16 and 48 GPU hours, respectively. We use NLTK for simple text processing such as lemmatization.
\section{Prompt Designs}\label{sec:sup_prompts}
\paragraph{Prompt for Silver-standard Dataset:} The prompt to generate silver-standard dataset is shown in Figure~\ref{fig:silver_dataset_prompt}. 
\paragraph{Prompt for Implicit Argument Prediction:} The prompt to perform implicit argument prediction using a cloze task is shown in Figure~\ref{fig:implicit_prompt}.
\paragraph{Prompt for Next Step Prediction:} The prompt to perform next step prediction is shown in Figure~\ref{fig:next_step_prompt}.
\section{Additional Analysis and Qualitative Results}\label{sec:sup_results}
\begin{figure*}[h!]
    \centering
    \scalebox{0.99}{%
    \begin{subfigure}[b]{0.45\textwidth}
        \includegraphics[width=\linewidth]{images/entities_count_what_n_iwhat_n.pdf}
        \caption{}
        \label{fig:supp_length_analysis_a}
    \end{subfigure}
    \hfill
    \begin{subfigure}[b]{0.45\textwidth}
        \includegraphics[width=\linewidth]{images/entities_count_where_n_iwhere_n.pdf}
        \caption{}
        \label{fig:supp_length_analysis_b}
    \end{subfigure}
    }

    \scalebox{0.99}{%
    \begin{subfigure}[b]{0.45\textwidth}
        \includegraphics[width=\linewidth]{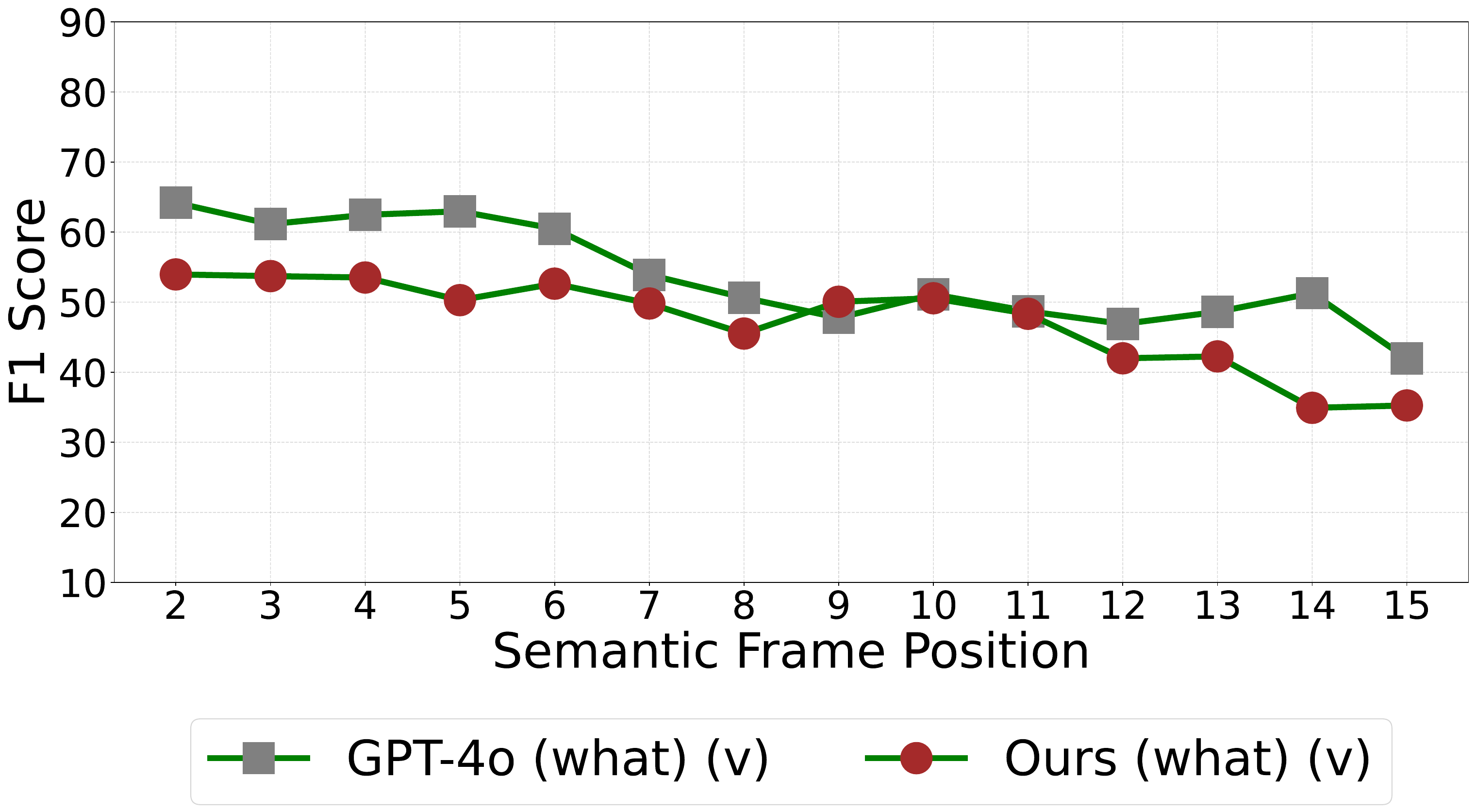}
        \caption{Only Video Input}
        \label{fig:supp_length_analysis_c}
    \end{subfigure}
    \hfill
    \begin{subfigure}[b]{0.45\textwidth}
        \includegraphics[width=\linewidth]{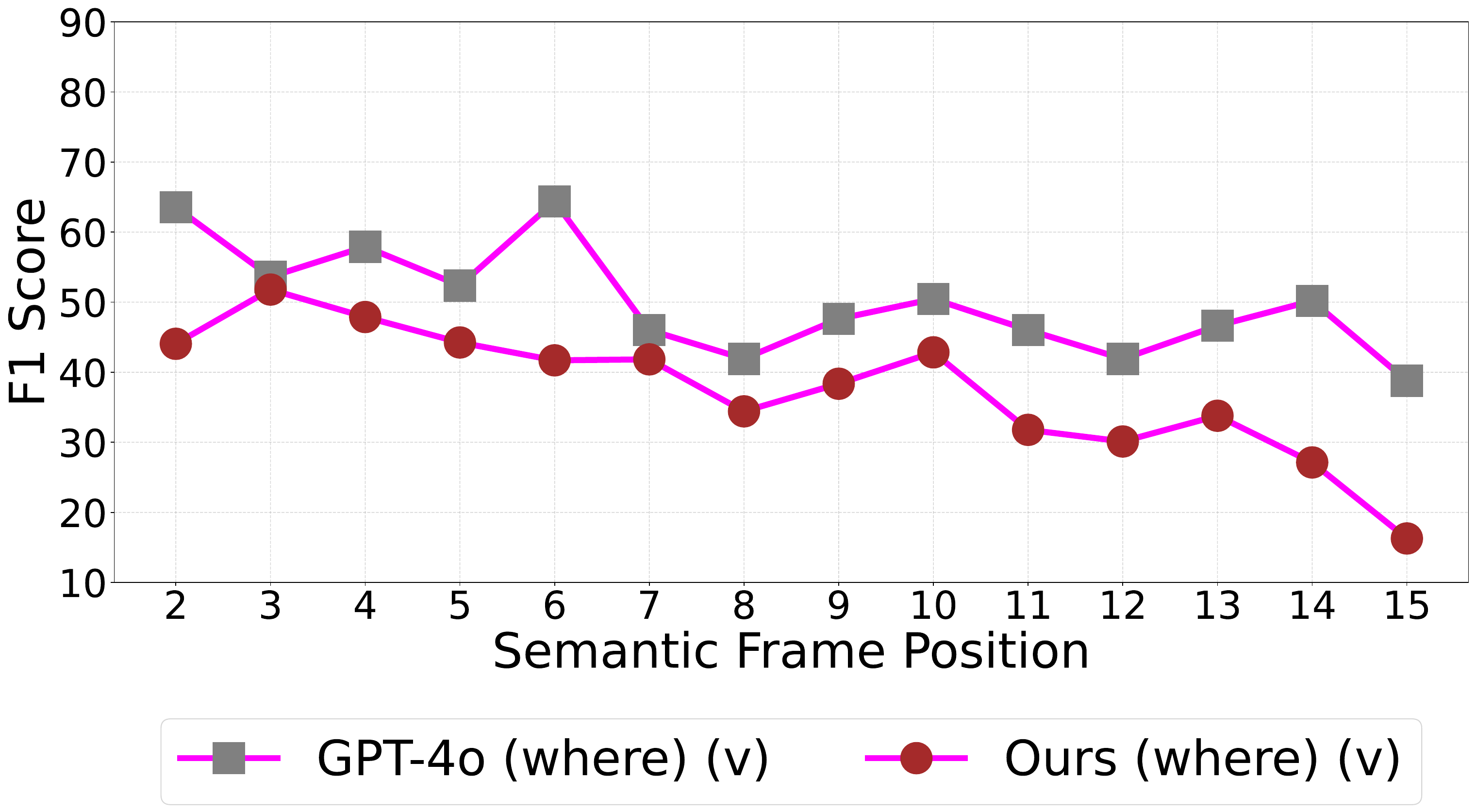}
        \caption{Only Video Input}
        \label{fig:supp_length_analysis_d}
    \end{subfigure}
    }

    \scalebox{0.99}{%
    \begin{subfigure}[b]{0.45\textwidth}
        \includegraphics[width=\linewidth]{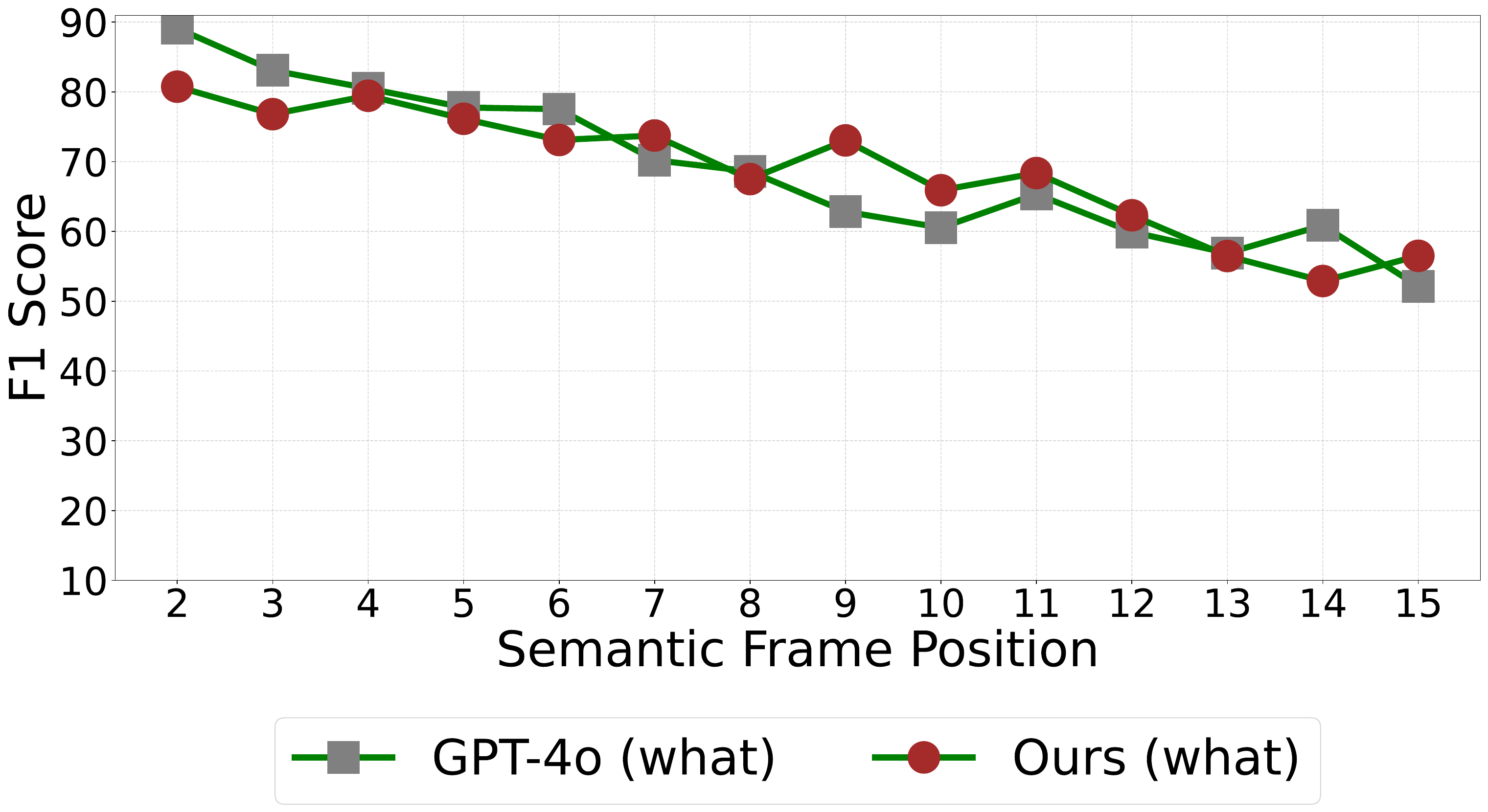}
        \caption{Video + Text Input}
        \label{fig:supp_length_analysis_c}
    \end{subfigure}
    \hfill
    \begin{subfigure}[b]{0.45\textwidth}
        \includegraphics[width=\linewidth]{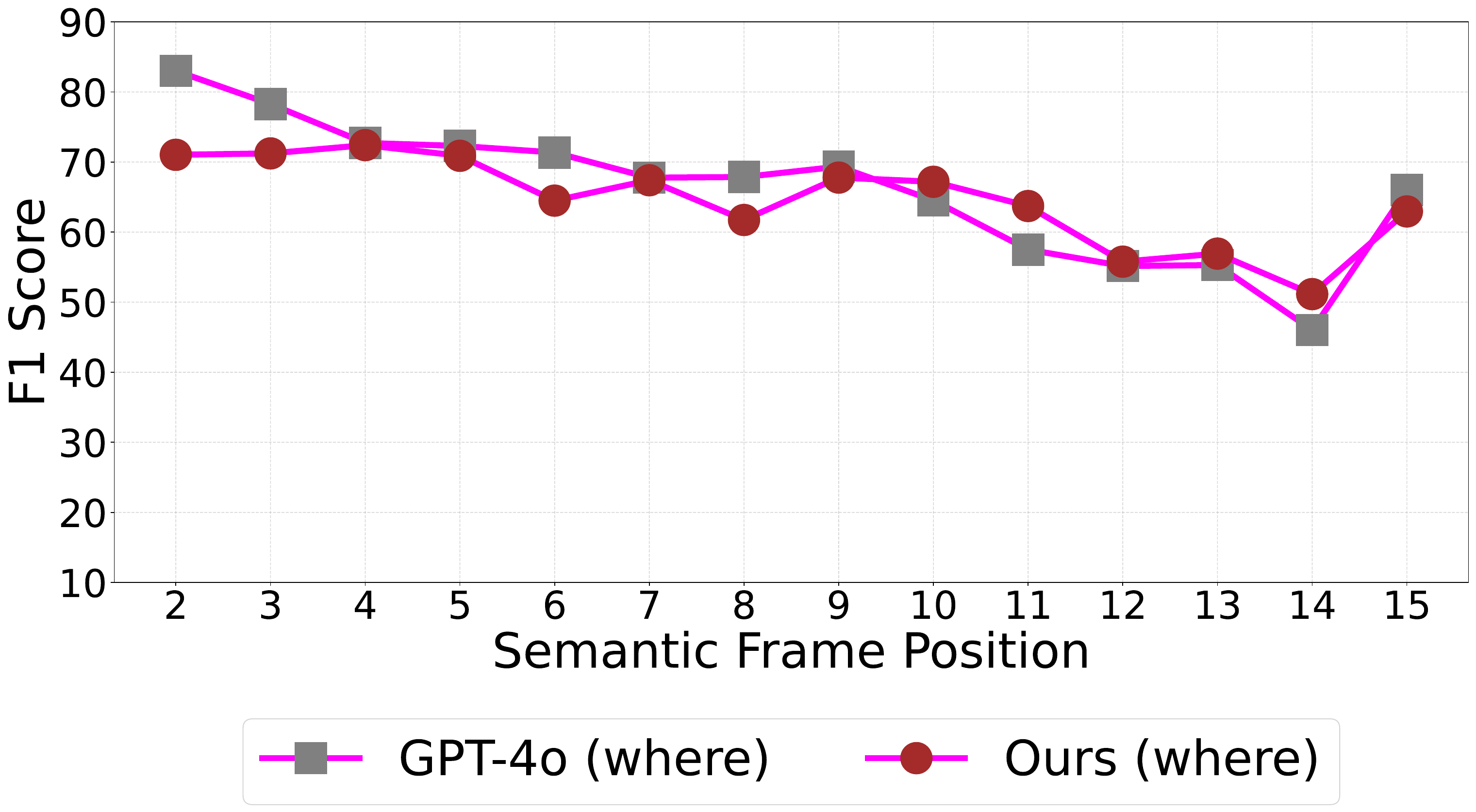}
        \caption{Video + Text Input}
        \label{fig:supp_length_analysis_d}
    \end{subfigure}
    }

    \scalebox{0.99}{%
    \begin{subfigure}[b]{0.45\textwidth}
        \includegraphics[width=\linewidth]{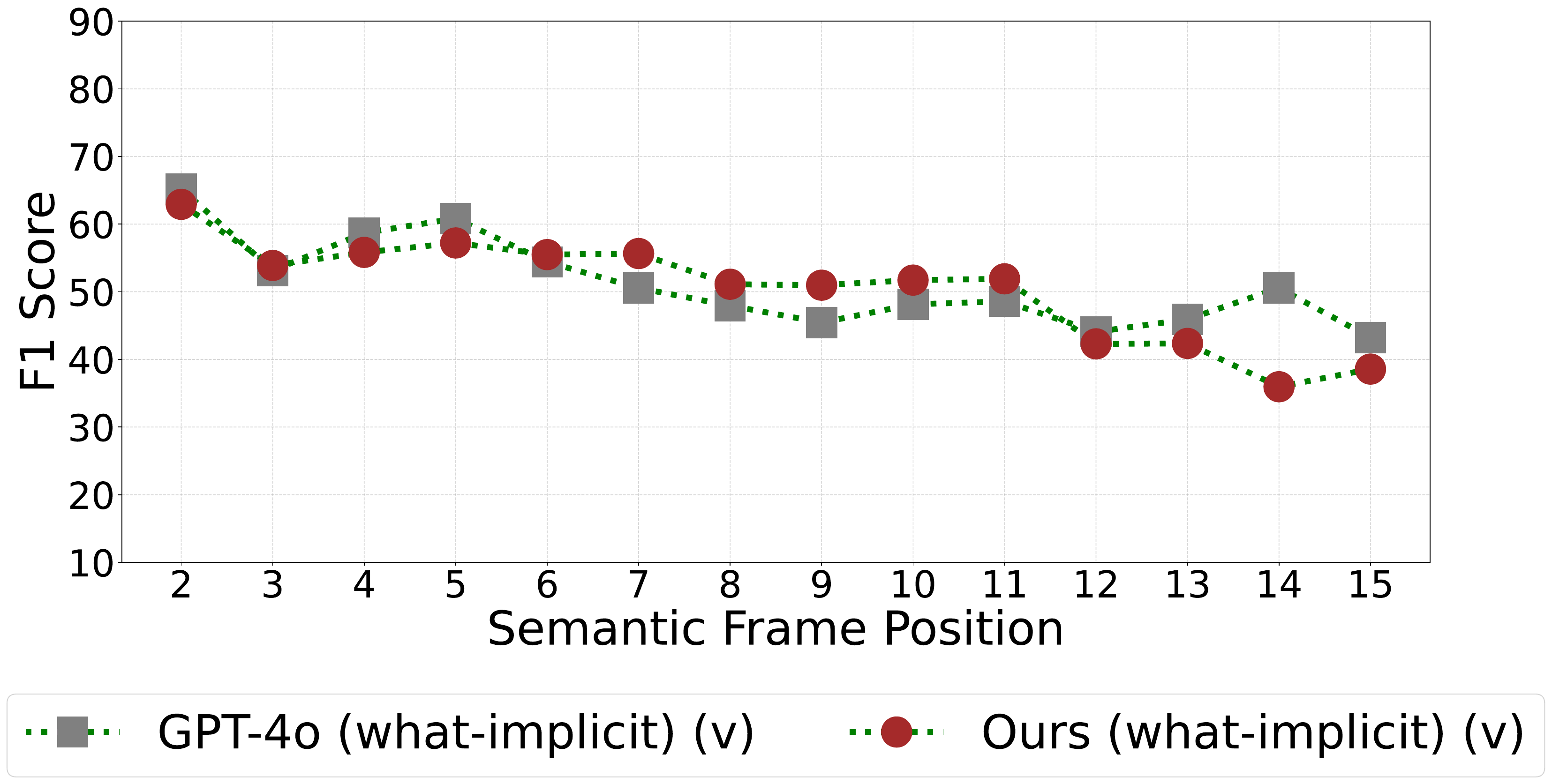}
        \caption{Only Video Input}
        \label{fig:supp_length_analysis_e}
    \end{subfigure}
    \hfill
    \begin{subfigure}[b]{0.45\textwidth}
        \includegraphics[width=\linewidth]{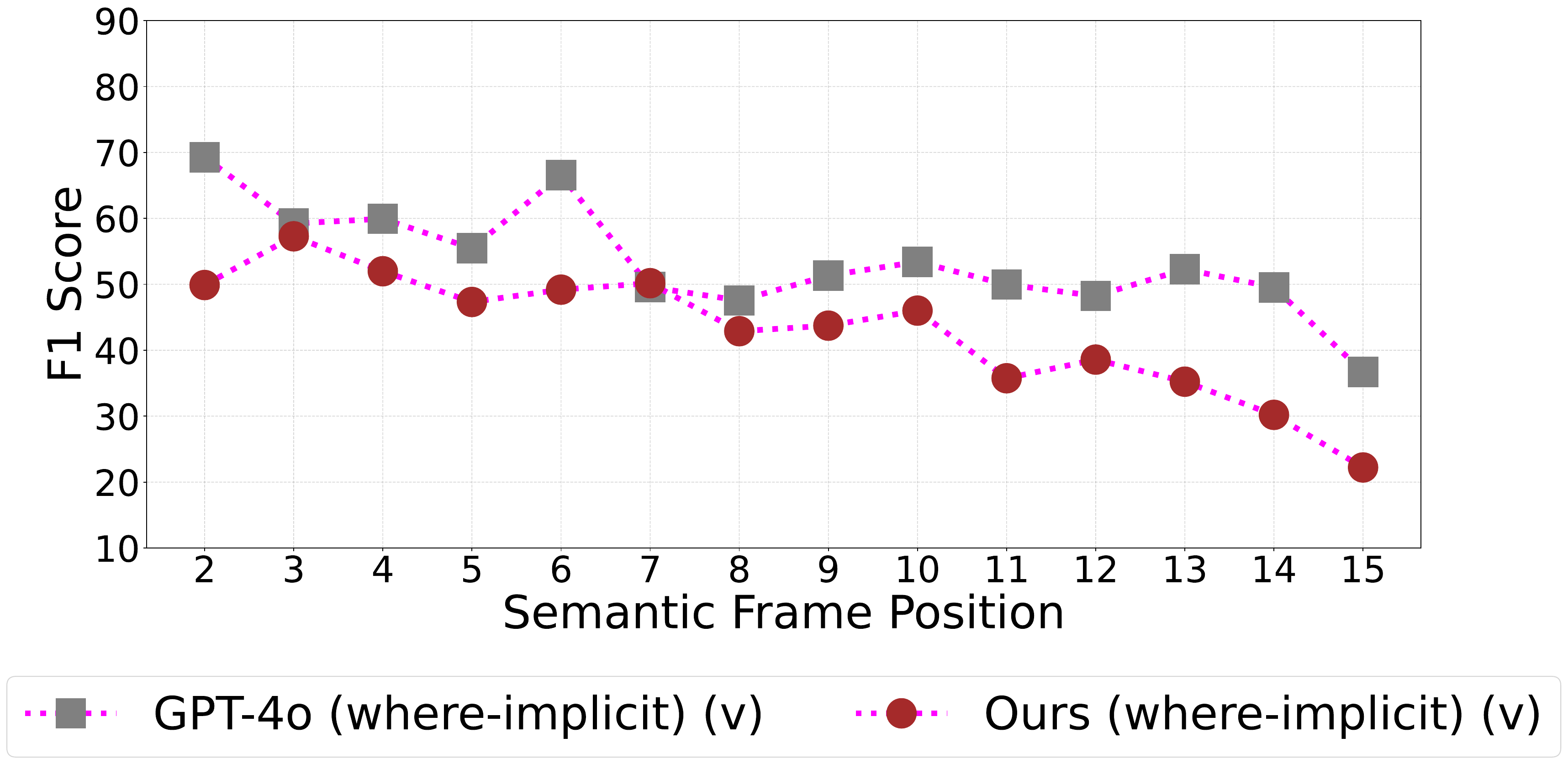}
        \caption{Only Video Input}
        \label{fig:supp_length_analysis_f}
    \end{subfigure}
    }
    \scalebox{0.99}{%
    \begin{subfigure}[b]{0.45\textwidth}
        \includegraphics[width=\linewidth]{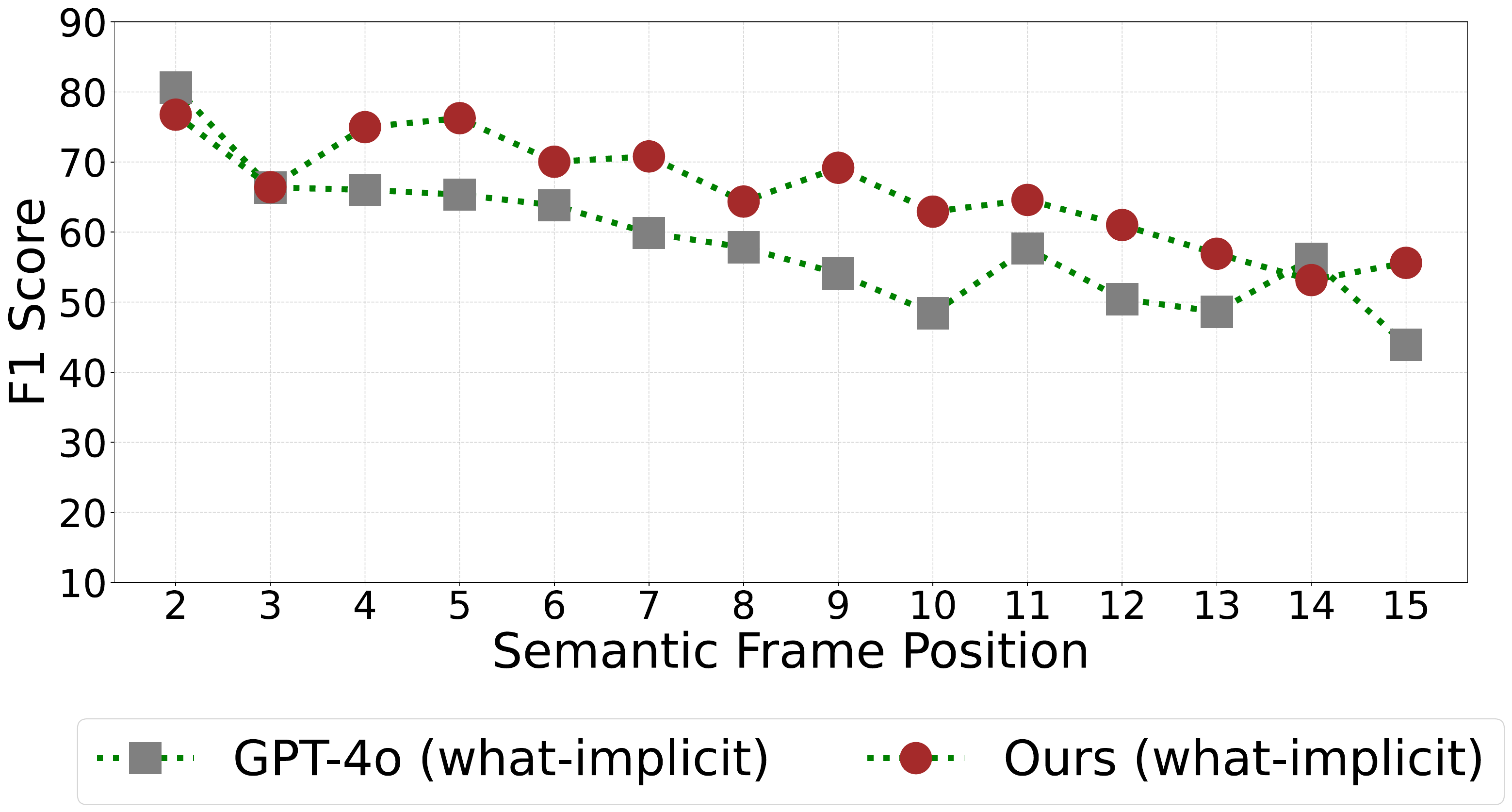}
        \caption{Video + Text Input}
        \label{fig:supp_length_analysis_e}
    \end{subfigure}
    \hfill
    \begin{subfigure}[b]{0.45\textwidth}
        \includegraphics[width=\linewidth]{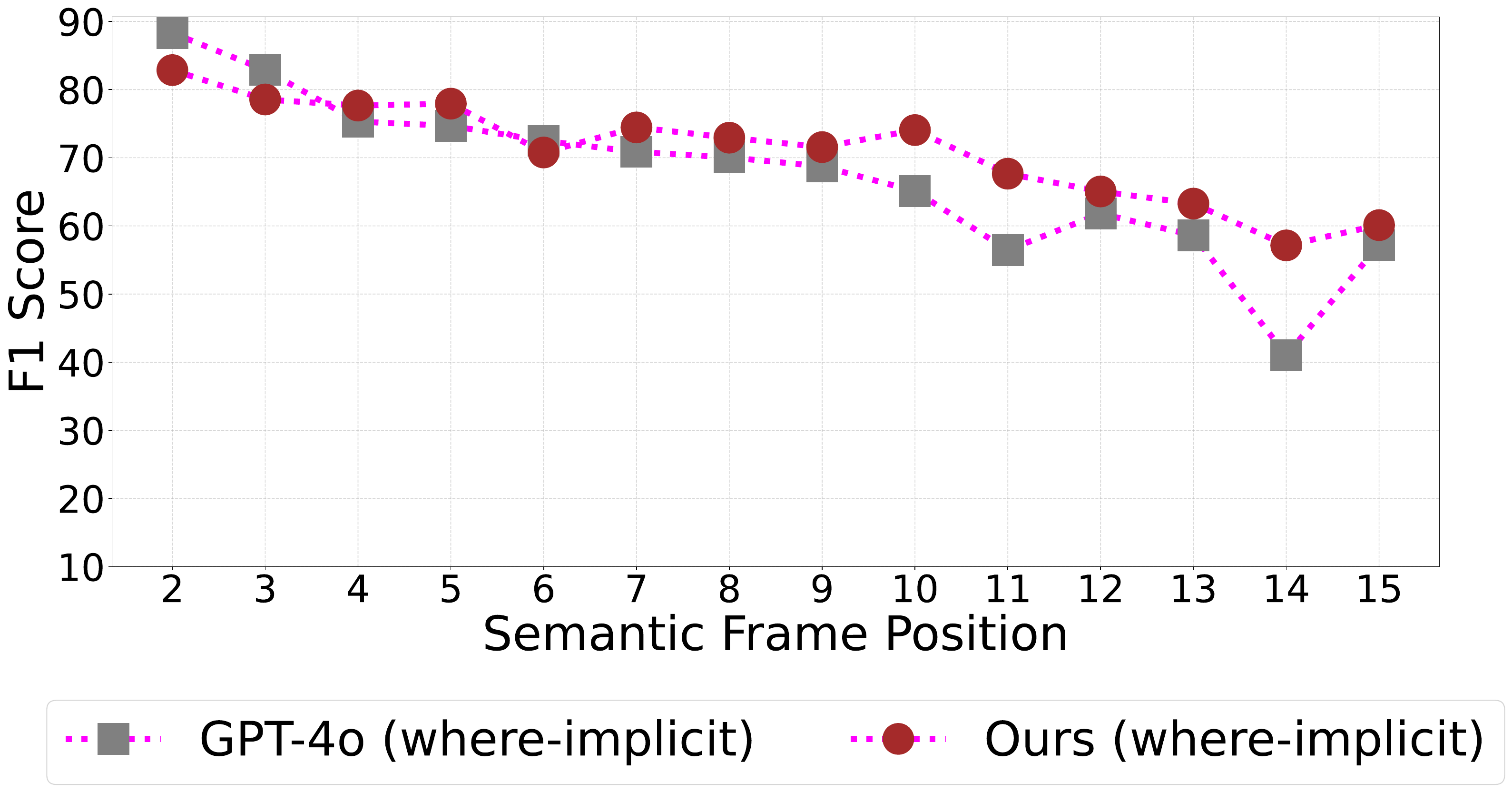}
        \caption{Video + Text Input}
        \label{fig:supp_length_analysis_f}
    \end{subfigure}
    }
    \caption{Comparison of GPT-4o and ours~\procsrlmodel~ models for argument prediction across \textit{semantic frame} positions.}
    \label{fig:supp_length_analysis}
\end{figure*}

\begin{figure*}[h!]
    \centering
    \scalebox{0.99}{%
    \begin{subfigure}[b]{0.95\textwidth}
        \includegraphics[width=\linewidth]{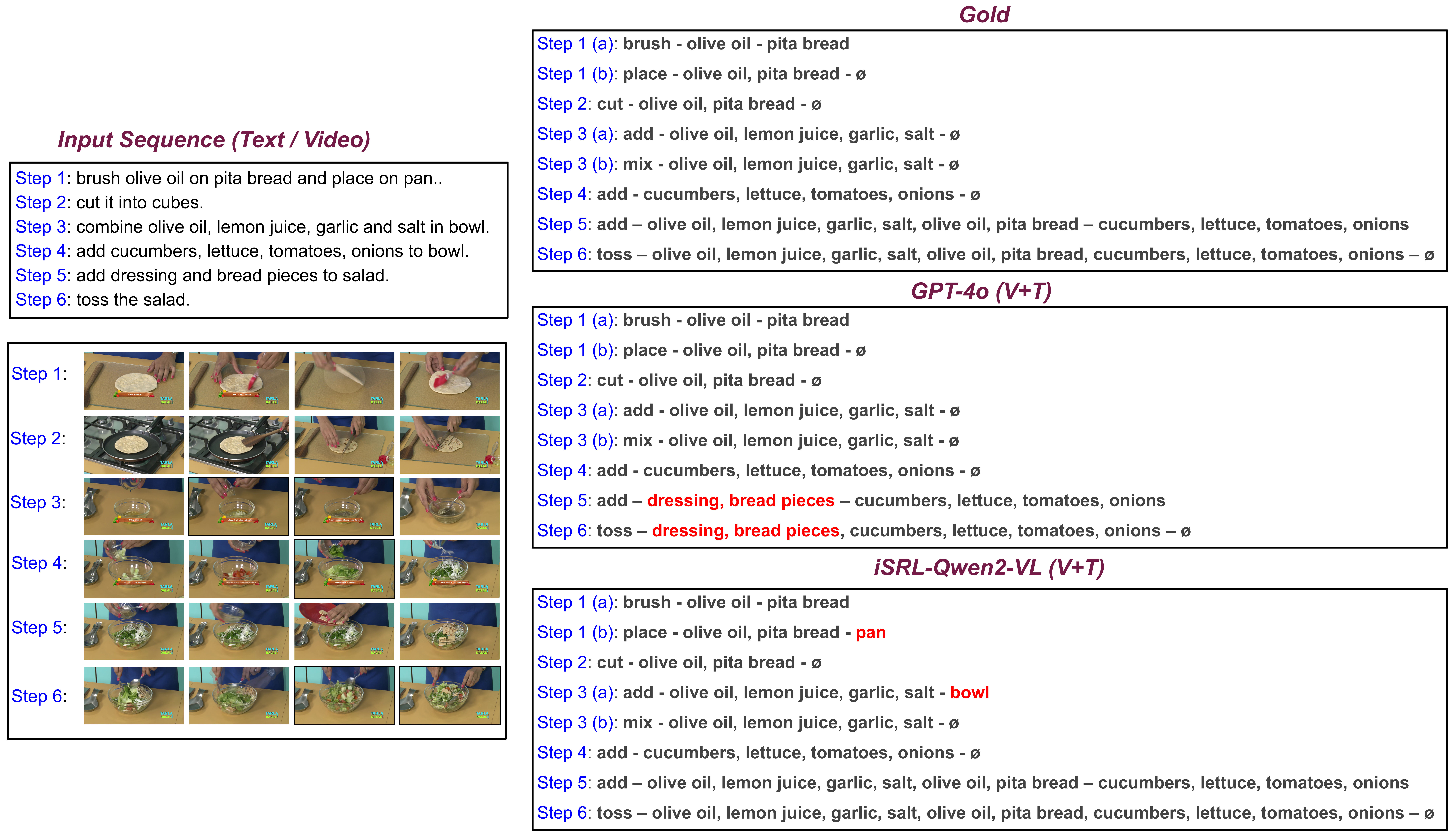}
        \caption{\textbf{Qualitative example using multimodal (\textbf{V}ideo + \textbf{T}ext) inputs.} The example is from Tasty~\cite{sener2022transferring} with ID-\href{https://tasty.co/recipe/cider-pulled-pork}{cider-pulled-pork}. The examples highlights that GPT-4o fail to identify \textit{dressing} as mixture output from step~3. Incorrect predictions are highlighted in \textcolor{red}{red}.}
        \vspace{2em}
        \label{fig:example2_result}
    \end{subfigure}
    }
    \scalebox{0.99}{%
    \begin{subfigure}[b]{0.95\textwidth}
        \includegraphics[width=\linewidth]{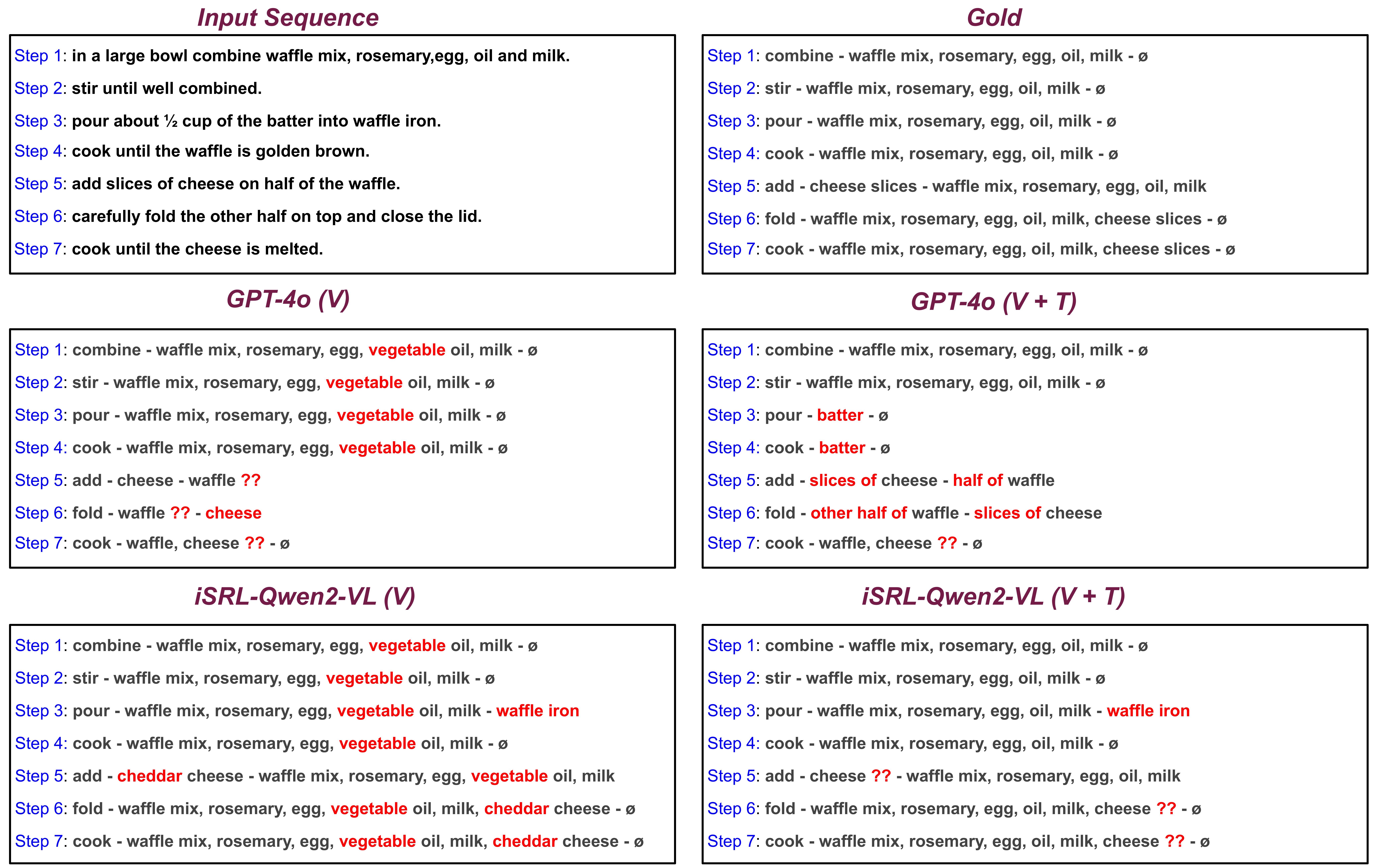}
        \caption{\textbf{Qualitative example using video (\textbf{V}ideo) and multimodal (\textbf{V}ideo + \textbf{T}ext) inputs.} Example is from Tasty~\cite{sener2022transferring} with ID-\href{https://tasty.co/recipe/waffle-grilled-cheese}{waffle-grilled-cheese}. Incorrect predictions are highlighted in \textcolor{red}{red}. The examples show that GPT-4o, with multimodal inputs (V+T), bias towards text, leading to incorrect predictions compared to video-only inputs.}
        \label{fig:example3_result}
    \end{subfigure}
    }
    \caption{Qualitative Examples from ~\procsrl~dataset.}
    \label{fig:more_example_result}
\end{figure*}

\paragraph{Fine-tuning Strategy} We ablate different components of Qwen2-VL model for fine-tuning with proposed silver-standard dataset generated by GPT-4o and show the results in Table~\ref{tab:pre_training}.
\begin{table}[h!]
    \centering
    \scalebox{0.8}{%
    \renewcommand{\tabcolsep}{0.4cm}
    \begin{tabular}{c|c|c|c|c}
        \hline
        \multirow{2}{*}{FT} & \multicolumn{2}{c}{SRL} & \multicolumn{2}{c}{SRL-Implicit} \\
        \cline{2-5}
            & $F1_{what}$ & $F1_{where}$ & $F1_{what}$ & $F1_{where}$ \\
        \hline
        ZS & 30.20 & 15.15 & 22.51 & 17.07 \\
        L & 46.07 & 33.72 & 45.59 & 36.38 \\
        V+L & 45.41 & 35.08 & 43.32 & 35.73 \\
        2-Stage & 46.21 & 33.43 & 45.76 & 36.06 \\
        \hline
    \end{tabular}
    }
    \caption{Ablation to study Fine-tuning~\procsrlmodel~ model with silver-standard dataset generated by GPT4o. \textbf{ZS}: zero-shot results for Qwen2-VL model. \textbf{L}: Only fine-tune the LLM. \textbf{V+L}: fine-tune visual encoder and LLM together. \textbf{2-Stage}: In first stage fine-tune only LLM and in next stage fine-tune only visual encoder of Qwen2-VL model.}
    \label{tab:pre_training}
\end{table}
\paragraph{Tracking Entities} We further extend our analysis for argument prediction across the semantic frame position in video only inputs in Figure~\ref{fig:supp_length_analysis} along with multimodal inputs.
\paragraph{Qualitative Results}
The additional qualitative results are shown in Figure~\ref{fig:more_example_result}.

\begin{figure*}[t]
    \centering
    % \captionsetup{font=footnotesize}
    \includegraphics[width=0.93\linewidth]{images/appendix/implicit_cloze_prompt.pdf}
    \caption{Prompt for Implicit Argument Prediction as Cloze task.}
    \label{fig:implicit_prompt}
\end{figure*}

\begin{figure*}[t]
    \centering
    % \captionsetup{font=footnotesize}
    \includegraphics[width=0.99\linewidth]{images/appendix/next_step_prompt.pdf}
    \caption{Prompt for Next Step Prediction with Semantic Frame.}
    \label{fig:next_step_prompt}
\end{figure*}

\begin{figure*}[t]
    \centering
    % \captionsetup{font=footnotesize}
    \includegraphics[width=0.99\linewidth]{images/appendix/silver_dataset_prompt.pdf}
    \caption{Prompt to generate Silver-standard dataset using GPT-4o~\cite{hurst2024gpt} model.}
    \label{fig:silver_dataset_prompt}
\end{figure*}

\end{document}